\documentclass[10pt,twocolumn,letterpaper]{article}

\usepackage{cvpr}              

\usepackage{graphicx}
\usepackage{amsmath}
\usepackage{amssymb}
\usepackage{booktabs}

\usepackage[accsupp]{axessibility}  

\usepackage[utf8]{inputenc} 
\usepackage[T1]{fontenc}    
\usepackage[pagebackref,breaklinks,colorlinks]{hyperref}  

\usepackage{url}            
\usepackage{booktabs}       
\usepackage{amsfonts}       
\usepackage{amsmath}       
\usepackage{nicefrac}       
\usepackage{microtype}      
\usepackage{xcolor}        
\usepackage{verbatim}        

\usepackage{algpseudocode} 
\usepackage{algorithm}

\makeatletter
\@namedef{ver@everyshi.sty}{}
\makeatother
\usepackage{tikz}

\definecolor{olivegreen}{rgb}{0, 0.6, 0}
\definecolor{black}{HTML}{000000}
\definecolor{white}{HTML}{ffffff}
\definecolor{color1}{HTML}{ACE5EE}
\definecolor{color2}{HTML}{0093AF}
\definecolor{color3}{HTML}{CC0000}
\definecolor{color4}{HTML}{0087BD}
\definecolor{color5}{HTML}{333399}
\definecolor{color6}{HTML}{20B2AA}

\usepackage{xspace}         
\usepackage{graphicx}

\usepackage{pgfplots}
\usepackage{pgfplotstable}
\usepgfplotslibrary{groupplots}

\usepackage{amsmath}
\usepackage{subcaption}

\usepackage{array, makecell} %
\newcolumntype{x}[1]{>{\centering\arraybackslash\hspace{0pt}}p{#1}}

\usepackage{subcaption}

\usepackage{multirow}

\usepackage{pifont}

\usepackage{ulem}

\newcommand{\aname}{AIT\xspace}
\newcommand{\lrname}{gradient inundation\xspace}
\newcommand{\LrName}{Gradient Inundation\xspace}
\newcommand{\Lrname}{Gradient inundation\xspace}
\newcommand{\lrabb}{GI\xspace}
\newcommand{\JL}[1]{{\color{olivegreen}[\textbf{\sc JLee}: \textit{#1}]}}
\newcommand{\KH}[1]{{\color{purple}[\textbf{\sc KH}: \textit{#1}]}}

\newcommand{\DK}[1]{{\color{teal}[\textbf{\sc DK}: \textit{#1}]}}

\newcommand{\new}[1]{{\color{violet}#1}}
\newcommand{\newdel}[1]{{\color{red}\sout{#1}}}

\renewcommand{\JL}[1]{}
\renewcommand{\KH}[1]{}
\renewcommand{\DK}[1]{}

\renewcommand{\new}[1]{{#1}}
\renewcommand{\newdel}[1]{}

\newcommand{\loss}{\mathcal{L}}

\def\blfootnote{\xdef\@thefnmark{}\@footnotetext}

\usepackage[resetlabels]{multibib}
\newcites{App}{References}

\usepackage[capitalize]{cleveref}
\crefname{section}{Sec.}{Secs.}
\Crefname{section}{Section}{Sections}
\Crefname{table}{Table}{Tables}
\crefname{table}{Tab.}{Tabs.}

\def\cvprPaperID{6131} 
\def\confName{CVPR}
\def\confYear{2022}

\begin{document}

\title{It's All In the Teacher: Zero-Shot Quantization Brought Closer to the Teacher}

\author{%
Kanghyun Choi\textsuperscript{1},
Hye Yoon Lee\textsuperscript{1},
Deokki Hong\textsuperscript{1},
Joonsang Yu\textsuperscript{2},\\%
Noseong Park\textsuperscript{1},
Youngsok Kim\textsuperscript{1},
and Jinho Lee\textsuperscript{1}\thanks{Corresponding author}\\%
\textsuperscript{1}College of Computing, Yonsei University\hspace{15pt}%
\textsuperscript{2}CLOVA ImageVision, CLOVA AI Lab, NAVER\\%
{\tt\small \textsuperscript{1}\{kanghyun.choi,hylee817,dk.hong,noseong,youngsok,leejinho\}@yonsei.ac.kr}\\%
{\tt\small \textsuperscript{2}joonsang.yu@navercorp.com}}
\maketitle

\begin{abstract}
  Model quantization is considered as a promising method to greatly reduce the resource requirements of deep neural networks.
To deal with the performance drop induced by quantization errors, 
a popular method is to use training data to fine-tune quantized networks.
In real-world environments, however, such a method is frequently infeasible because training data is unavailable due to security, privacy, or confidentiality concerns.
Zero-shot quantization addresses such problems, usually by taking information from the weights of a full-precision teacher network to compensate the performance drop of the quantized networks. In this paper, we first analyze the loss surface of state-of-the-art zero-shot quantization techniques and provide several findings.
In contrast to usual knowledge distillation problems, 
zero-shot quantization often suffers from 1) the difficulty of optimizing multiple loss terms together, and 2) the poor generalization capability due to the use of synthetic samples.
Furthermore, we observe that many weights fail to cross the rounding threshold during training the quantized networks even when it is necessary to do so for better performance.
Based on the observations, we propose AIT, a simple yet powerful technique for zero-shot quantization, which addresses the aforementioned two problems in the following way: AIT i) uses a KL distance loss only without a cross-entropy loss, and ii) manipulates gradients to guarantee that a certain portion of weights are properly updated after crossing the rounding thresholds.
Experiments show that AIT outperforms the performance of many existing methods by a great margin, taking over the overall state-of-the-art position in the field.

\end{abstract}

 \section{Introduction}
Deep neural network quantization~\cite{binaryconnect,hwang2014fixed,dorefa, lin2016fixed} is a powerful tool for improving the computational efficiency of deep neural networks (DNNs).
When being accompanied with the \newdel{right}\new{low-bitwidth} hardware design\newdel{(of low-bitwidth)}~\cite{outliernpu, bitfusion, tpu}, \newdel{both }the latency and energy consumption of \newdel{deep neural networks}\new{DNNs} can be greatly reduced.

One problem of quantized models is, however, that they often suffer from the significant drop in accuracy, mainly due to quantization errors~\cite{lin2016fixed}.
A popular way to address the problem is to further train or calibrate the model with training data~\cite{choi2018pact,jacob2018quantization,rastegari2016xnor,zhang2018lq,zhou2017incremental,dorefa}. 
During the fine-tuning procedure, the forward pass is performed with quantized values whereas the backpropagation is done with floating-point values to recover the accuracy loss in the initial quantization.

Unfortunately, such fine-tuning methods, which assume the full availability of training data at the time of quantization, are often not feasible in reality. 
Many models are disclosed to public only with their trained weights, and the dataset may contain proprietary, confidential, or sensitive data that fundamentally prohibit sharing.

\JL{too long. get to the point}
\emph{Zero-shot quantization} (or data-free quantization)~\cite{outlier,aciq,dfq,choi2020data,dsg,gdfq,zeroq,qimera,autorecon} is therefore a necessary technique for quantization.
It assumes that only the architecture and the pre-trained weights are available at the time of quantization.
Current successful approaches are mainly led by generative approaches~\cite{zeroq,gdfq,zaq,dsg,autorecon,qimera}. 
Using synthetic samples from generators, knowledge distillation~\cite{hinton2015distilling} is applied against full-precision models.
It is known that the state-of-the-art methodology achieves almost similar performance to that of the data-driven approaches (i.e., quantization with real samples) for 5-bit fixed-point quantization, and comparable performance on 4-bit fixed-point setting.~\cite{qimera}.

\JL{hey, what happened to CE label smoothing? sensitivity on CE? (maybe that has some meaning in cifar100?)}

However, the recipe of the fine-tuning in zero-shot quantization is mainly adopted from common knowledge distillation problems~\cite{hinton2015distilling,haroush2020knowledge,cho2019efficacy} that consider neither quantization nor synthetic samples.
As in the knowledge distillation, the loss function of the zero-shot quantization is \emph{habitually} built as a combination of the cross-entropy (CE) against the hard label and the Kullback–Leibler (KL) divergence against the full-precision network's output.\footnote{In the remainder of this paper, we refer to \emph{CE} as the cross-entropy against the hard label and \emph{KL} as the KL divergence against the full-precision network unless otherwise stated. \JL{would this CE/KL definitions be okay?}}
%
It works well in practice, but there \newdel{@@do not exist any explicit }\new{are no detailed }studies to introspect the appropriateness of the loss in the context of zero-shot quantization. 
\newdel{Therefore, there is an urgent need for more analyses on whether those solutions are appropriate.}
\new{Therefore, more analyses on those solutions are needed.}
Moreover, the distribution of synthetic samples can be different from that of the original data.
In such a case, they can be considered a type of adversarial samples (also see \cref{fig:samples} for examples) and thus, the quantized network produces a huge generalization gap.
%

To our knowledge, 
we for the first time perform in-depth analyses on the loss surface of the zero-shot quantization problem.
Through the analyses, we find several key observations for better quantization.
First, quantized models often have difficulty optimizing multiple loss terms, and the loss terms fail to cooperate --- in other words, the angle between the gradients of CE and KL is quite large in many cases.
Second, KL usually has a much flatter loss surface than that of CE, having a better potential for generalizability.

To this end, we propose a method to address such problems of the zero-shot quantization, called \textit{\aname (All In the Teacher)}.
While pursuing a flatter surface of the loss curve, \aname lets the quantized student model get closer to the full-precision teacher model. 
To be more specific, we exclude CE from the loss, and apply our proposed \emph{\lrname} with KL only.
In addition, \lrname is designed to grow the gradients of KL in such a manner that a certain portion of weights are guaranteed to be updated in each layer.
As a result, the quantized model approaches closer to the full-precision teacher, and our method takes over the state-of-the-art position for various datasets.
Our contributions can be summarized as follows: \JL{are there too many?}
\begin{itemize}
    \item We analyze the first and second-order loss surfaces, i.e., \newdel{Jacobian}\new{gradient} and Hessian, of the zero-shot quantization problem. To the best of our knowledge, we are the first to closely investigate the loss function in the zero-shot quantization problem.
    
    \item We identify that the gradients from CE and KL form a large angle from the beginning to the end of the fine-tuning. This implies that the quantized network is \newdel{making a trade-off between the two instead of benefiting from them}\new{suffering from their trade-off instead of benefiting from them working in harmony}.
 
    \item We analyze the local curvature of the loss surface and observe that the two losses of our interest exhibit a great amount of curvature difference. 
    
    \item We observe that the quantized student suffers from infrequent updates, where only a few layers are changing their integer weights and the remaining layers are stuck below rounding thresholds.
    
    \item Based on these findings, we propose \aname which excludes the cross-entropy loss, and manipulates the gradients using our proposed \lrname method such that the quantized student model can faithfully resemble the full-precision teacher model. 
    
    \item We perform a thorough evaluation of \aname. The results show that \aname outperforms the existing algorithms by a great margin, showing the state-of-the-art performance on the zero-shot quantization problem.
\end{itemize}

\section{Background and Related Work}
\subsection{Quantization}
Quantization of neural networks have been studied for a while, and there are numerous methods~\cite{han2015deep, park2017weighted, cai2017deep, choi2017towards, faraone2018syq, jung2019learning}.
In this work, we consider symmetric, uniform quantization that is known to be much easier to build hardware architectures for.
With $n$ bits, a weight parameter $\theta$ is represented by one of $2^n$ ranges.
We use $\theta^q$ to denote the quantized weights as outputs of a quantization function $Quant()$. 
For $Quant()$, we use a simple yet efficient function as the following \cite{jacob2018quantization}:
\begin{align}
    \theta^q&=Quant(\theta) = \lfloor \theta \times S - z \rceil,\\
    S &= \frac{2^n -1}{\theta_{max}-\theta_{min}}, \\
    z &= S \times \theta_{min} + 2^{n-1},
\end{align}
where $S$ is the scaling factor to convert the range of $\theta$ to $n$ bit, and $z$ decides which quantized value zero is mapped to.
After quantization, the quantized integer value represents $\theta'\in\mathbb{R}$ obtained by dequantization:
\begin{align}
    \theta' = \frac{(\theta^q+z)}{S}.
    \label{eq:quant_fake}
\end{align}
The procedure is the same for activation values, except that the minimum and the maximum are obtained from observing activations from a few batches and taking a moving average.

\JL{KH: chk. can i change this to $\theta^q$?}

\subsection{Zero-shot Quantization}
\label{sec:zq}
Even though quantization has been shown to be effective even for extremely low bits~\cite{binaryconnect, dorefa, liu2018bi, rastegari2016xnor}, they usually require training data to fine-tuning or calibration.
Zero-shot quantization is a method to relax the privacy or confidentiality problem of the training data.

Earlier methods for zero-shot quantization were focused on how to build a good quantization function $Quant()$, by using schemes such as weight equalization, bias correction, or range adjustments~\cite{aciq,outlier,dfq}.
Among them, ZeroQ~\cite{zeroq} was the first work to introduce the notion of distilled data that is designed to match the batch-norm stats of the original full-precision network.
\newdel{@@}With this scheme, choosing the adequate mixed-precision quantization for each layer has been proposed together.
On top of ZeroQ, DSG~\cite{dsg} added diverse sample generation to improve the performance.

Later, GDFQ~\cite{gdfq} adopted generative models~\cite{cgan,acgan} to create better samples.
The generator $G$ and the quantized model $Q$ are jointly trained with the following loss functions:
\begin{align}
    \loss_{GDFQ}(G) &= (1-\alpha)\loss_{CE}^P(G) + \alpha\loss_{BNS}^P(G),
    \label{eq:gdfq_G} \\
    \loss_{GDFQ}(Q) &= (1-\delta) \loss_{CE}^Q(Q) + \delta\loss_{KL}^P(Q),
    \label{eq:gdfq_Q}
\end{align}
where both utilizes cross-entropy ($L_{CE}$), while the generator matches the batch normalization stats from the full-precision model ($L_{BNS}$) and the quantized network optimizes KL divergence ($L_{KL}$).
Variants of GDFQ are currently forming state-of-the-art family of the zero-shot quantization, by adopting better generator~\cite{autorecon}, adversarial training~\cite{zaq}, or boundary-supporting sample generations~\cite{qimera}.
In this work, we provide an in-depth analysis of the loss function $\loss_{GDFQ}(Q)$, and a novel scheme to improve its performance.


\begin{figure}
\centering
\centering
\includegraphics[width=0.9\columnwidth]{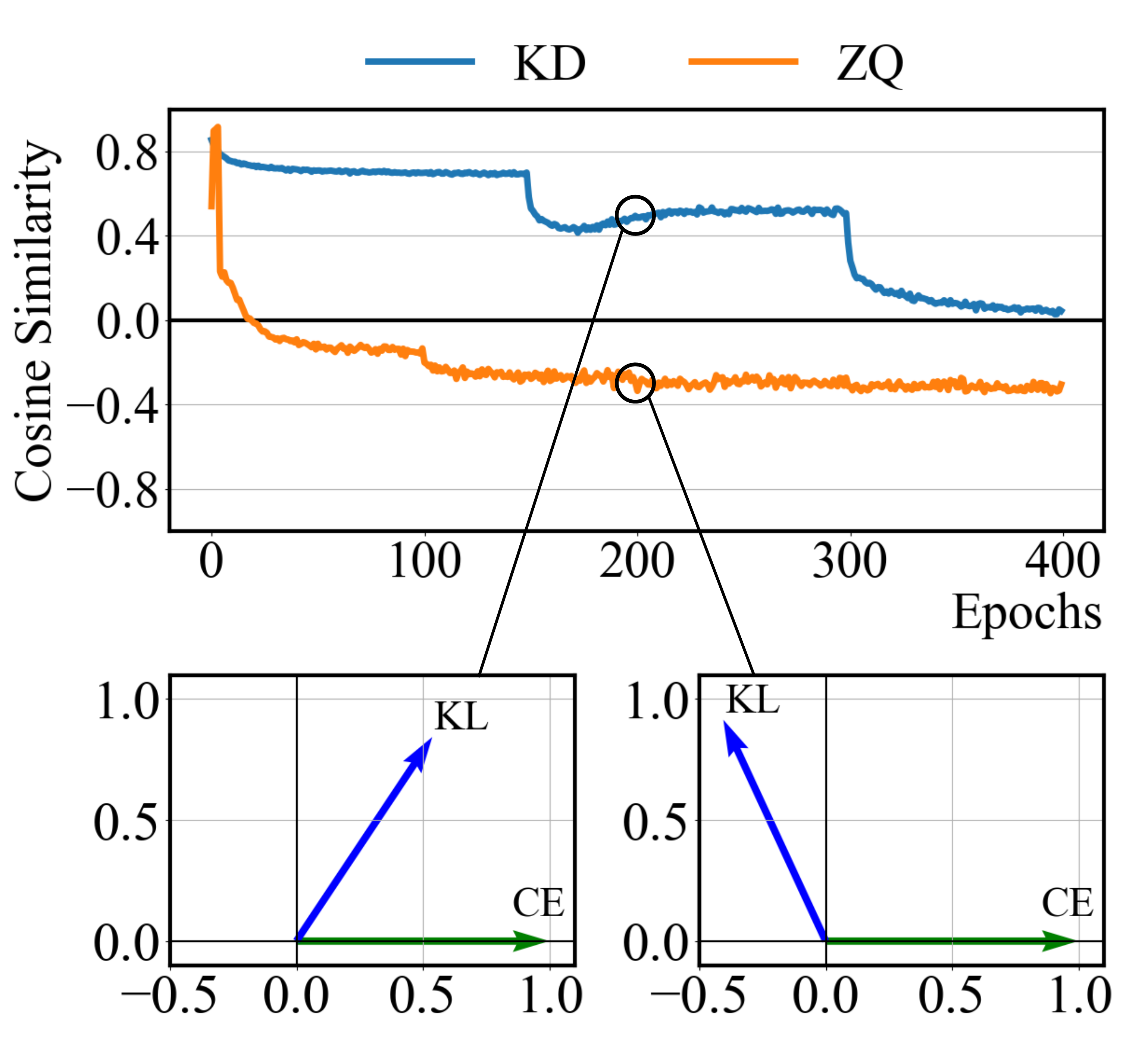}
\vspace{-2mm}
\caption{Plot of cosine similarity between cross-entropy on hard labels KL divergence on the full-precision model. At the bottom are snapshots of the relative angle between KL and CE of knowledge distillation (KD, bottom left) and zero-shot quantization (ZQ, bottom right) at epoch 200.}
\vspace{-3mm}
\label{fig:cos}
\end{figure}

\newcommand{\subfigwidth}{0.45\textwidth}

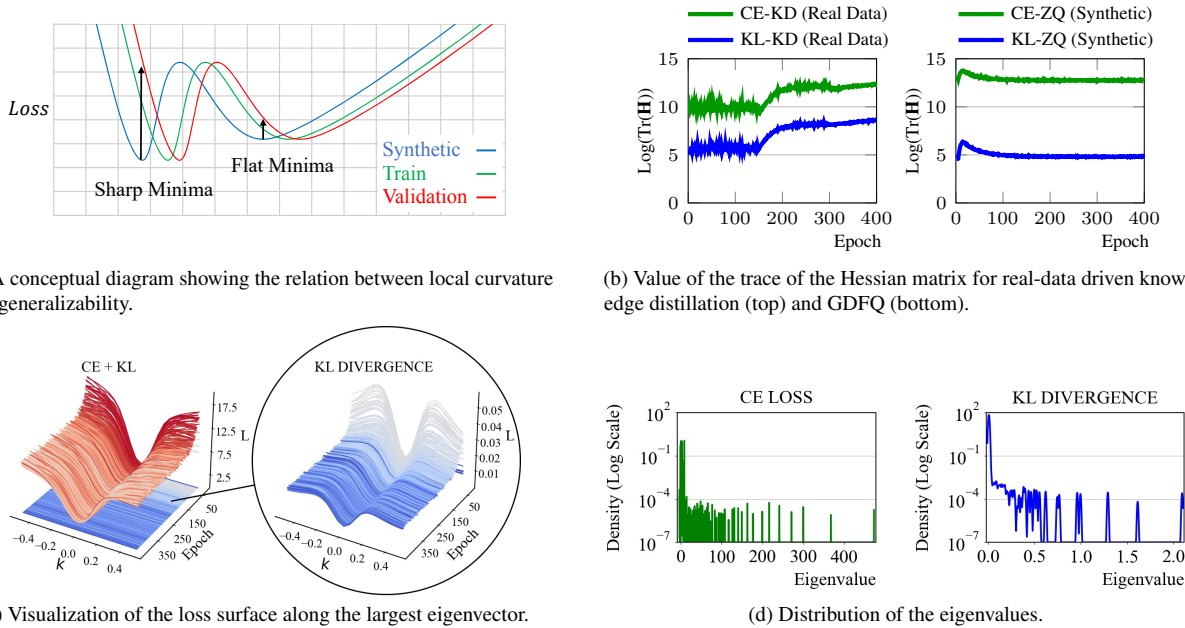
\begin{figure*}[t]
\centering
\begin{subfigure}[t]{0.45\textwidth}
\centering
\includegraphics[width=0.9\textwidth]{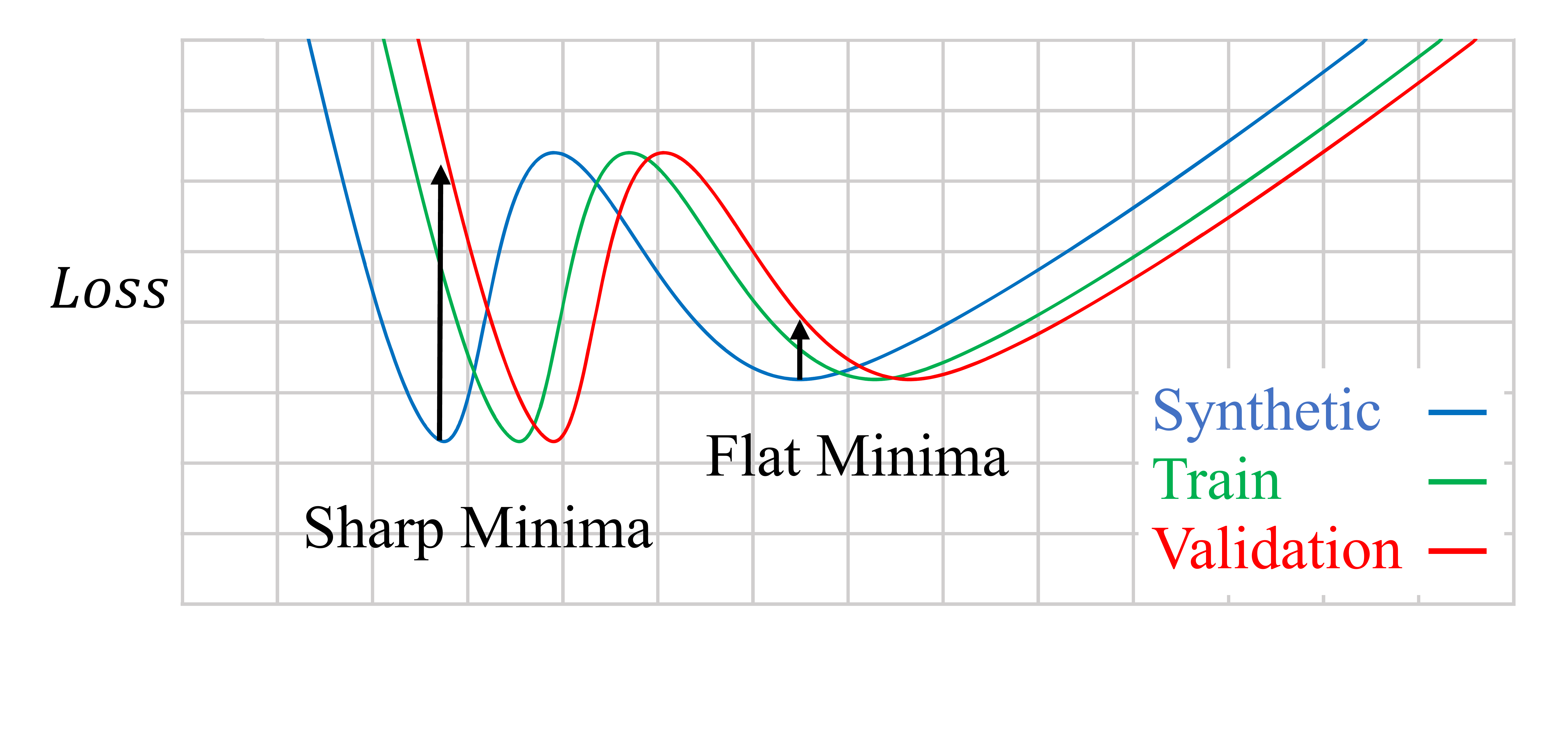}
\caption{A conceptual diagram showing the relation between local curvature and generalizability.}
\label{fig:concept}
\end{subfigure}
\hspace{5mm}
\begin{subfigure}[t]{0.45\textwidth}
    \begin{tikzpicture}

\centering
\hspace{3mm}
\begin{groupplot}[group style={vertical sep=2.8em,horizontal sep=3em,group size= 2 by 1},height=3.6cm,width=0.5\columnwidth]
\nextgroupplot[
width=.52\textwidth, 
height=3.5cm,
xmajorticks=true,
ymajorgrids,
xmin=0,
xmax=400,
ymax=15,
ymin = 0,
xlabel = {Epoch},
xlabel shift=-.3em,
xlabel near ticks,
ylabel near ticks,
xlabel style={at={(ticklabel cs:0.9)},font =\rmfamily\scriptsize},
x tick label style = {font =\scriptsize, text width = 1.4cm, align = center, anchor = north},
ylabel={Log(Tr(\textbf{H}))},
ylabel style = {font=\scriptsize, yshift=-0.2cm},
y tick label style = {font =\scriptsize, anchor = east, 
 /pgf/number format/fixed, 
},
legend cell align={left},
legend style={draw=none, fill=none, at={(0.54,1.0), font=\scriptsize},
anchor=south,legend columns=1,
/tikz/every even column/.append style={column sep=0.1cm}},
]  

\addplot[mark=none,ultra thick,color=olivegreen] table [x=epoch, y=CEKD10] {figs/hessian/hessiandata.txt};\addlegendentry{CE-KD (Real Data)}
\addplot[mark=none,ultra thick,color=blue] table [x=epoch, y=KLKD10] {figs/hessian/hessiandata.txt};\addlegendentry{KL-KD (Real Data)}

\nextgroupplot[
width=.52\textwidth, 
height=3.5cm,
xmajorticks=true,
ymajorgrids,
xmin=0,
xmax=400,
ymax=15,
ymin = 0,
xlabel = {Epoch},
xlabel shift=-.3em,
xlabel near ticks,
ylabel near ticks,
xlabel style={at={(ticklabel cs:0.9)},font =\rmfamily\scriptsize},
x tick label style = {font =\scriptsize, text width = 1.4cm, align = center, anchor = north},
ylabel={Log(Tr(\textbf{H}))},
ylabel style = {font=\scriptsize, yshift=-0.2cm},
y tick label style = {font =\scriptsize, anchor = east, 
 /pgf/number format/fixed, 
},
legend cell align={left},
legend style={draw=none, fill=none, at={(0.54,1.0), font=\scriptsize},
anchor=south,legend columns=1,
/tikz/every even column/.append style={column sep=0.1cm}},
]  

\addplot[mark=none,ultra thick,color=olivegreen] table [x=epoch, y=CEGDFQ] {figs/hessian/hessiandata.txt};\addlegendentry{CE-ZQ (Synthetic)}
\addplot[mark=none,ultra thick,color=blue] table [x=epoch, y=KLGDFQ] {figs/hessian/hessiandata.txt};\addlegendentry{KL-ZQ (Synthetic)}

\end{groupplot}

\end{tikzpicture}  
\caption{Value of the trace of the Hessian matrix for real-data driven knowledge distillation (top) and GDFQ (bottom).\JL{to 400 epochs}}
\label{fig:hessian}
\end{subfigure}
\begin{subfigure}[t]{0.45\textwidth}
\centering
\includegraphics[width=0.9\textwidth]{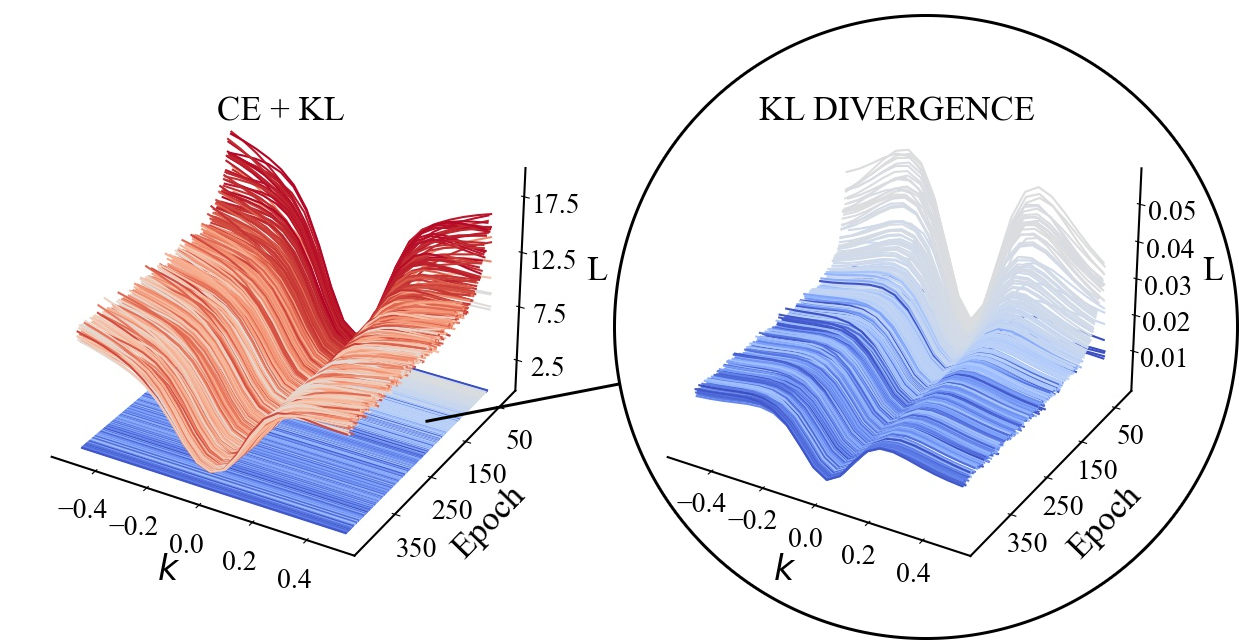}
\caption{Visualization of the loss surface along the largest eigenvector.}
\label{fig:visual}
\end{subfigure}
\hspace{5mm}
\begin{subfigure}[t]{0.45\textwidth}
\centering
\includegraphics[width=1.0\textwidth]{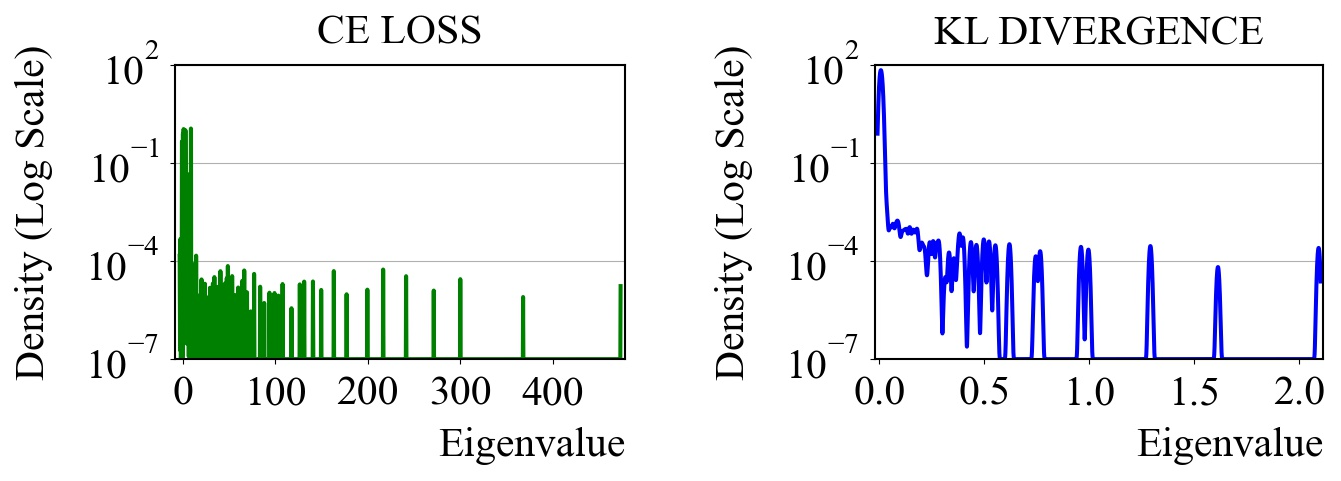}
\caption{Distribution of the eigenvalues.}
\label{fig:eigenvalue}
\end{subfigure}
\vspace{-1mm}
\caption{Analysis of the loss surface.}
\vspace{-3mm}
\label{fig:surface}
\end{figure*}

\section{Analyses on the Zero-shot Quantization}
\label{sec:diagnosis}
In this section, we provide in-depth analyses on the first/second-order loss surface of the state-of-the-art zero-shot quantization method.
We studied the impact of CE and KL on the loss function of zero-shot quantization setting, and reveal that they are not cooperative but hinder each other (\cref{sec:cos}).
Then, we investigate the difference of the local curvature with the lens of Hessian.
Because the zero-shot quantization suffers from larger generalization gaps, finding the solution of the flatter minima is critical (\cref{sec:generalization}).

\subsection{Gradient Cosine Similarity}
\label{sec:cos}

In this subsection, we attempt to find a partial answer to the question: \emph{are CE and KL cooperative in quantization?} \JL{'zero shot' temporarily taken off}
As discussed in \cref{sec:zq}, loss functions from current techniques for zero-shot quantization~\cite{gdfq,autorecon,qimera} are mainly composed of the CE against the hard labels, and KL divergence against the full-precision teacher.

\JL{put Quant cause before synthetic}
However, it has been discussed that better models do not necessarily make good teachers when the student has limited capacity.
In such cases, the student often has to make trade-off between KL and CE~\cite{cho2019efficacy}.
As quantized models have much lower representational capability~\cite{nahshan2019loss, jacob2018quantization}, it could be difficult for them to optimize both terms. 
\JL{toned down a bit}
Furthermore, with synthetic samples not exactly matching the distribution of the real samples,
the labels associated to each sample via hard labels and teacher outputs can be distinct, contributing to the difficulty of addressing both losses.

In such regard, \cite{auxloss} suggested using cosine similarity of the two gradients as a metric for determining whether an auxiliary can contribute toward a single main task.
The authors suggest that the two losses should be used together only in steps where their cosine similarity is larger than zero (when they form an acute angle).


Thus, inspired by the proposals of \cite{auxloss}, we analyzed the training of the zero-shot quantization as the following. 
Using GDFQ~\cite{gdfq} as a representative for zero-shot quantization, we measured the cosine similarity of the $g_{KL}$ and $g_{CE}$ while training a quantized ResNet-20 model with synthetic data.
The full-precision teacher is pre-trained, and the generator is jointly trained with the quantized student using \cref{eq:gdfq_G} and \cref{eq:gdfq_Q} with \newdel{$\alpha=0.5$ and  $\delta=0.5$}\new{$\alpha=\delta=0.5$}, respectively.
For comparison, we have also measured the same metric from a common real data (CIFAR-10) based knowledge distillation with pre-trained ResNet-20 as a teacher and random initialized ResNet-20 as a student (self-distillation) using the same loss function.

The results are presented in \cref{fig:cos}. 
Using the real samples denoted as `KD', the KL and CE terms induce gradients of the similar direction, supporting the common wisdom that the combination of them works well in practice.
However, with the zero-shot quantization setting using synthetic samples denoted as `ZQ', the cosine distance between them take negative values.
The bottom two plots visualize the angle of the gradients.
In both plots, the \newdel{CE gradient vector }\new{$g_{CE}$} is set to (1,0), and \newdel{KL gradient }\new{$g_{KL}$} is plotted to preserve the relative angle to the \newdel{CE}\new{$g_{CE}$}. 

The trend persists throughout the training, as shown in the plot.
Right after the training begins,
the cosine distance of ZQ becomes negative and it is maintained until the end of training, while that of the KD is positive.
This implies that combinations of the two losses do not cooperate well with each other, and using them together could potentially harm the model performance.
Although we display only one case for clarity, 
the same trend was observed across many models and datasets. 
Refer to the Appendix for further results.

\subsection{Generalizability}
\label{sec:generalization}

\JL{chk this later}
The observation in \cref{sec:cos} suggest that using only one of the two losses --- KL divergence against the full-precision teacher or CE against the hard label --- could be better for the problem.
Some work~\cite{cho2019efficacy} suggest modifying the teacher for distillation, but such method is unavailable in a zero-shot setting \newdel{because we only have access to the weights, and not the training data of the full-precision teacher.}\new{because we have no access to the training data}

In such regard, we examine the generalizability of the \newdel{two }loss terms\newdel{ of our interest}.
Including zero-shot quantization, diverse applications relying on synthetic samples~\cite{park2018data,shankarampeta2021few,choi2021dual,van2021class,liu2020generative,yin2020dreaming,lee2021invertible} usually suffer from huge generalization gap, coming from the discrepancy in the data distributions.
One can easily infer that quantization requires stronger generalization when performed under a zero-shot environment.

To evaluate the generalizability, we measure local curvature of the loss surface. 
Popularly measured with the Hessian matrix $\bold{H}$ ($\frac{\partial^2 L}{\partial \theta^2} \in \mathbb{R}^{n\times n}$, where $\theta$ is a vector of $n$ weight parameters), local curvature of the loss surface is a metric that is drawing much attention from the field, and it is thought to hold a key to better generalization~\cite{keskar2016large,jastrzebski2020break,jastrzkebski2018relation,jastrzkebski2017three,chaudhari2019entropy, achille2017critical, jastrzebski2021catastrophic}.
As illustrated in \cref{fig:concept}, if the optimizer settles at a sharp minima, the performance at test time is likely to incur a larger degradation compared that of a flat minima. 
Such gap would be much larger with synthetic data incorrectly modeling the validation data distribution.
In line with this finding, \newdel{there are numerous pieces of literature that supports}\new{many literatures support} the claim of smaller local curvature improving generalization~\cite{keskar2016large,jastrzebski2020break,jastrzkebski2018relation,jastrzkebski2017three,chaudhari2019entropy, achille2017critical, jastrzebski2021catastrophic}.

%
%
%
%

\newcommand{\subfigurewidth}{0.3\textwidth}
\begin{figure*}[t]
\centering
\begin{subfigure}[t]{\subfigurewidth}
    \begin{tikzpicture}

\centering
\begin{groupplot}[group style={vertical sep=2.8em,horizontal sep=3em,group size= 1 by 1},height=3.8cm,width=0.58\columnwidth]

\nextgroupplot[
width=\textwidth, 
height=4cm,
xmajorticks=true,
ymajorgrids,
xmin=0,
xmax=400,
ymax=1.5,
xlabel = {Epoch},
xlabel shift=-.3em,
xlabel near ticks,
ylabel near ticks,
xlabel style={at={(ticklabel cs:0.9)},font =\rmfamily\scriptsize},
x tick label style = {font =\scriptsize, text width = 1.4cm, align = center, anchor = north},
ylabel={Inter-epoch Cosine Similarity},
ylabel style = {font=\scriptsize, yshift=0cm},
y tick label style = {font =\scriptsize, anchor = east, 
 /pgf/number format/fixed, 
},
legend cell align={left},
legend style={draw=none, fill=none, at={(0.54,1.0), font=\scriptsize},
anchor=south,legend columns=3,
/tikz/every even column/.append style={column sep=0.1cm}},
]  

\addplot[mark=none,color=orange] table [x=epoch, y=GDFQ] {figs/disparity/disparitydata.txt};\addlegendentry{ZQ}
\addplot[mark=none,color=olivegreen] table [x=epoch, y=KD] {figs/disparity/disparitydata.txt};\addlegendentry{KD (Real Data)}

\end{groupplot}

\end{tikzpicture}  
 \vspace{-4mm}
\caption{Cosine similarity of the gradients between epochs.\label{fig:cossim}}
\end{subfigure}
\hspace{3mm}
\begin{subfigure}[t]{\subfigurewidth}
\centering
\includegraphics[width=\textwidth, trim={0 1.5cm 0 0},clip=true]{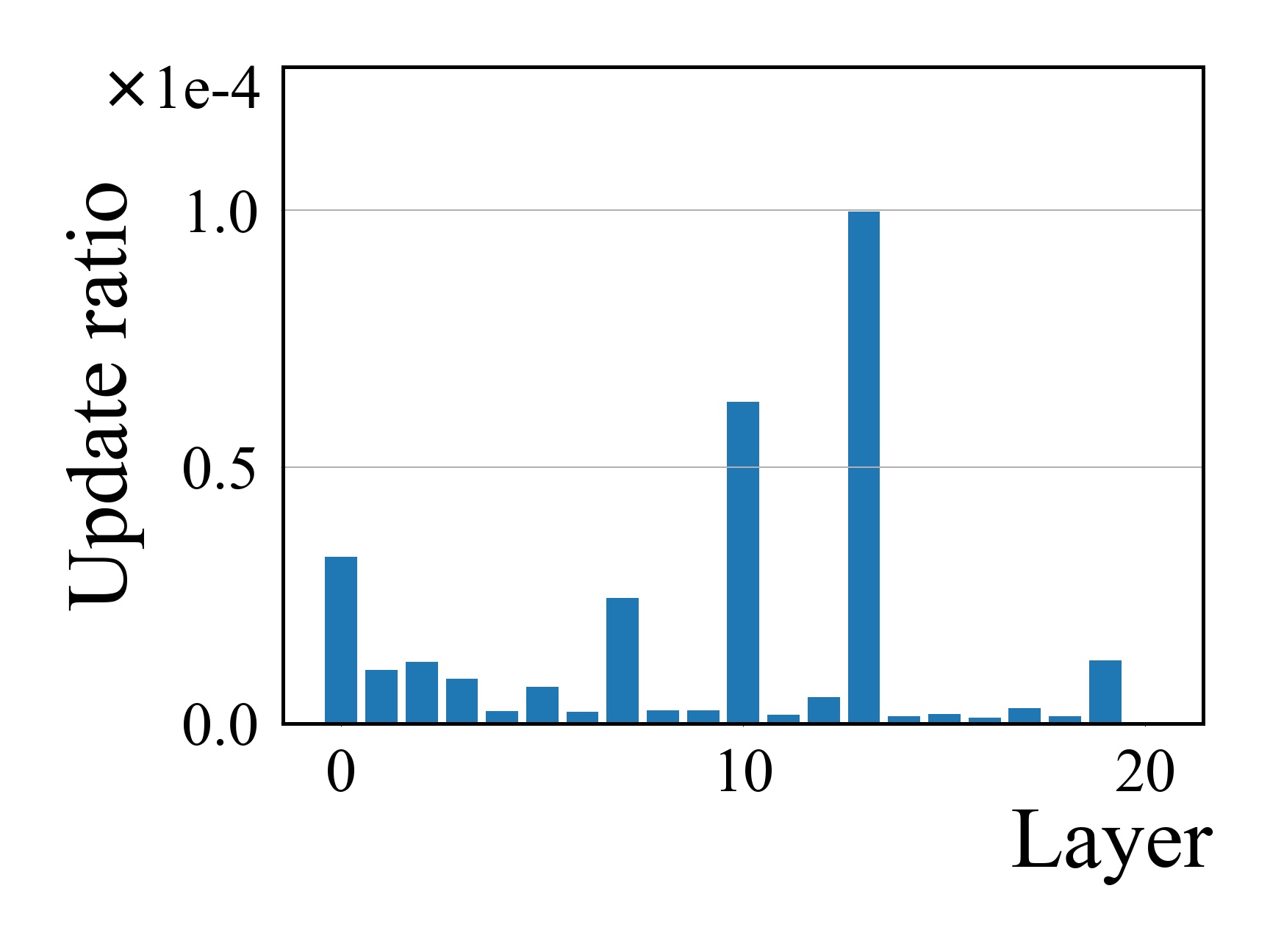}
\caption{Number of quantized parameters updated per each layer of ResNet-18 at epoch 60. \label{fig:count_before}}
\end{subfigure}
\hspace{3mm}
\begin{subfigure}[t]{\subfigurewidth}
\centering
\includegraphics[width=\textwidth, trim={0 1.7cm 0 0},clip=true]{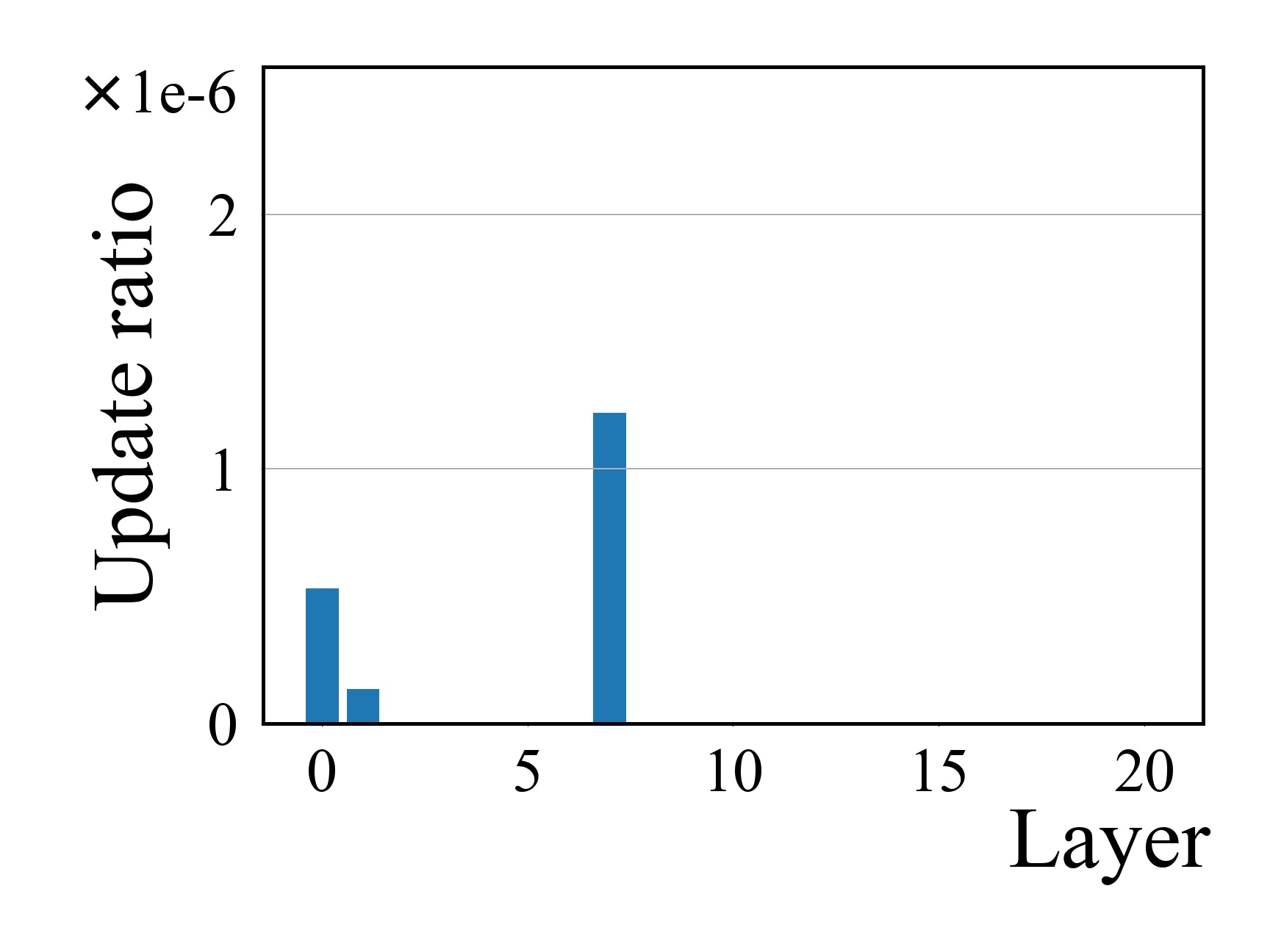}
\caption{Number of quantized parameters updated per each layer ResNet-18 at epoch 350.\label{fig:count_after}}
\end{subfigure}
\vspace{-1mm}
\caption{Analysis on the naive KL-only zero-shot quantization.}
\vspace{-3mm}
\label{fig:disparity}
\end{figure*}
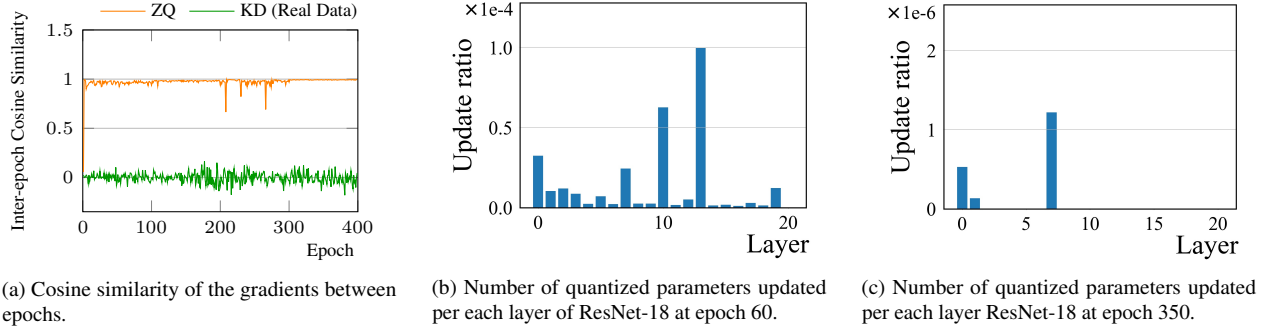

\cref{fig:hessian} plots $Tr(\bold{H})$, the trace of the Hessian matrix, approximated by PyHessian~\cite{yao2020pyhessian} implementing Lanczos algorithm~\cite{lanczos1950iteration}.
We separate Hessian calculation for each of CE and KL.
The trace values are significantly different, where that of the KL \newdel{divergence }is much smaller than the CE.
The gap is notably larger in zero-shot quantization (right, \emph{ZQ}) than on real-data knowledge distillation (left, \emph{KD}).
In addition, the distribution of the eigenvalues displayed in \cref{fig:eigenvalue} also show a huge difference in the local curvature of the \newdel{two }loss terms.
\JL{KH: chk the following, too}
While CE has longer tail for high eigenvalues, those of KL has more concentration to lower \new{eigen}values\newdel{ with a higher peak near zero}.
\JL{chk later}

This could potentially lead to two conclusions: the loss surface of the KL divergence is much flatter, or the model has converged to the minima in the loss surface of the KL divergence.
However, in our case, we believe it advocates for the former\newdel{ of flatter loss surface}, based on an auxiliary experiment.
It is commonly observed that near the minima, the disparity between gradients start to arise~\cite{thiran2020early, mahsereci2017early}.
Following the same regard, 
we measure the cosine distance of the gradient averaged within an epoch, compared to that of the previous epoch (\emph{inter-epoch cosine similarity}).
As displayed in \cref{fig:cossim}, the gradients of KL in the zero-shot quantization setting (ZQ) points to a consistent direction (large cosine similarity) compared to that of the real data distillation (KD),
indicating that it has not reached the minima yet.

\cref{fig:visual} illustrates a more direct visualization of the loss surface.
From the Hessian matrix, we took the largest eigenvector $e$ at each epoch, we plotted the value of CE and KL on the left plot by calculating $L(\theta(t) + k \cdot e \cdot \hat{g}(t))$ for $k \in [-0.5,0.5]$, where 
$\hat{g}$ is the average gradient along $e$.
The left plot presents the CE in red color scheme and KL in blue color scheme. 
It is clearly shown that the surface is flatter in the KL surface especially near the end of training\newdel{ (epoch 400)}. 


\subsection{Summary}
\JL{remove this one?}
Summarizing the studies in this section,
we first observed that the CE and KL form a large angle in the gradient space, 
and the quantized model has difficulty optimizing both directions.
Furthermore, by measuring the stats from Hessian matrices, 
we conclude that KL has a much flatter loss surface for potentially better generalization, which is an important issue for generative zero-shot quantization methods.


\section{AIT Method}
In this section, we describe our \emph{AIT (All In the Teacher)} method in detail.
From the observations in \cref{sec:diagnosis}, we first drop CE term from the loss, and apply a novel \emph{\lrname} to bring the quantized model as close as possible to the full-precision teacher.

\subsection{KL-only Zero-shot Quantization}
\label{sec:kl}
Motivated by the experiments from \cref{sec:generalization}, we ran GDFQ~\cite{gdfq} with KL-only loss (i.e., $\delta=1$).
However, as will be shown later in \cref{tab:ablation}, the performance severely degrades in all settings. \KH{check degradation.}
We find an explanation from the experiments of \cref{fig:cossim}.
Even toward the end of the training, the direction of the \newdel{gradients from the KL divergence}\new{$g_{KL}$} remains consistent, and training for more epochs did not solve the problem.
This indicates that the model did not converge at the minima of the KL surface.


Another set of experiments shown in \cref{fig:count_before,fig:count_after} gives a closer look at the phenomenon.
We count the average number of weight parameters that cross the rounding threshold (parameters whose quantized values have changed from the previous step).
We make two observations:
First, the portion of quantized values crossing the rounding threshold is extremely small. 
Even when training has not stabilized (epoch 60), only 0.0011\% of weights are being updated each step. 
At a later epoch (350), the phenomenon becomes worse that only four values are updated in a whole epoch, which is only 1.8e-7\% of weight updates per step during the epoch. 
In addition, the changes are extremely unbalanced, 
where all the updates are only occurring in just three layers.

We posit that this is from the quantized training process that constrains integer value updates. 
During the training, the quantized network internally stores its full-precision values.
The parameters are quantized for the forward pass of the backpropagation, and the gradients are applied to the internal full precision values.
As the gradient values become smaller after a few epochs of training, the change in the parameters are usually not large enough to cross the threshold, and only a few layers are continuously making changes,
stopping the model from moving towards a lower point in the loss surface.

\begin{table*}[]
    \centering
    \footnotesize

    \resizebox{0.99\textwidth}{!}
    {
    \begin{tabular}{x{1.5cm}|c|c|c|cc|cc|cc}
    \toprule
    Dataset & \makecell{Model\\(FP32 Acc.)} & Bits & ZeroQ & 
    GDFQ & \makecell{GDFQ\\{\footnotesize+\aname}}  &
    Qimera & \makecell{Qimera\\{\footnotesize+\aname}} & 
    ARC & \makecell{ARC\\{\footnotesize+\aname}}  
    \\
    \midrule

	\multirow{2}{*}{CIFAR-10}	&	ResNet-20	&	4w4a	&	79.30	&	90.25	&	91.23	( 	$+$0.98	)	&	\textbf{91.26}	&	91.23	( 	$-$0.03	)	&	88.55	&	90.49	( 	$+$1.94	)			\\	
		&	93.89	&	5w5a	&	91.34	&	93.38	&	93.41	( 	$+$0.03	)	&	\textbf{93.46}	&	93.43	( 	$-$0.03	)	&	92.88	&	92.89	( 	$+$0.01	)			\\	\midrule
	\multirow{2}{*}{CIFAR-100}	&	ResNet-20	&	4w4a	&	47.45	&	63.39	&	\textbf{65.80}	( 	$+$2.41	)	&	65.10	&	65.40	( 	$+$0.30	)	&	62.76	&	61.05	( 	$-$1.71	)			\\	
		&	70.33	&	5w5a	&	65.61	&	66.12	&	\textbf{69.26}	( 	$+$3.14	)	&	69.02	&	\textbf{69.26}	( 	$+$0.24	)	&	68.40	&	68.40	( 	$+$0.00	)			\\	\midrule
	\multirow{6}{*}{ImageNet}	&	ResNet-18	&	4w4a	&	22.58	&	60.60	&	65.51	( 	$+$4.91	)	&	63.84	&	\textbf{66.83}	( 	$+$2.99	)	&	61.32	&	65.73	( 	$+$4.41	)			\\	
		&	71.47	&	5w5a	&	59.26	&	68.40	&	70.01	( 	$+$1.61	)	&	69.29	&	69.22	( 	$-$0.07	)	&	68.88	&	\textbf{70.28}	( 	$+$1.40	)			\\	
		&	ResNet-50	&	4w4a	&	8.38	&	52.12	&	64.24	( 	$+$12.12	)	&	66.25	&	67.63	( 	$+$1.38	)	&	64.37	&	\textbf{68.27}	( 	$+$3.90	)			\\	
		&	77.73	&	5w5a	&	48.12	&	71.89	&	74.23	( 	$+$2.34	)	&	75.32	&	75.54	( 	$+$0.22	)	&	74.13	&	\textbf{76.00}	( 	$+$1.87	)			\\	
		&	MobileNetV2	&	4w4a	&	10.96	&	59.43	&	65.39	( 	$+$5.96	)	&	61.62	&	66.81	( 	$+$5.19	)	&	60.13	&	\textbf{66.47}	( 	$+$6.34	)			\\	
		&	73.03	&	5w5a	&	59.88	&	68.11	&	71.70	( 	$+$3.59	)	&	70.45	&	71.68	( 	$+$1.23	)	&	68.40	&	\textbf{71.96}	( 	$+$3.56	)			\\		
	
    \bottomrule

    \end{tabular}}
    \vspace{-1mm}
        \caption{Comparison on \aname with data-free quantization schemes. \JL{two digits fo decimals!}}
        \vspace{-3mm}
    \label{tab:master}
\end{table*}

\subsection{\LrName}

To address the problem of KL-only method, we propose \lrname (\lrabb).
Overall, we attempt to dynamically manipulate the gradients \new{$g_l$} of each layer \new{$l$}, such that certain number of parameters are guaranteed to update in its integer value.
\new{With stochastic gradient descent, consider the update rule of parameter $\theta_{l,k}$ at step $k$ with learning rate $\eta$:}
\begin{align}
    \theta_{l,k+1} &= \theta_{l,k} - \eta \cdot g_{l,k}.
\end{align}
with \lrname, the modified rule is as the following:
For the parameters $\theta_{l,k}$, quantized parameters $\theta_{l,k}^q$ and the corresponding gradients $g_{l,k}$ from layer $l$, 
\begin{align}
    \theta_{l,k+1} &= \theta_{l,k} - \eta \cdot g'_{l,k}, \\
    g'_{l,k} &= \kappa_{l} \cdot g_{l,k}, \\
    \kappa_{l} &= \text{arg} \min\limits_{\kappa_l} \lVert \Delta \theta^q_{l,k} - T \rVert, \label{eq:GI_objective} \\
    \Delta \theta^q_{l,k} &= \sum \mathbb{I}(\theta^q_{l,k} \neq \theta^q_{l,k+1}), \\
    T &= \rho \cdot dim(\theta_l),
\end{align}
where $\rho \in [0,1]$ is a predetermined proportion that exceeds the quantization threshold, $\mathbb{I()}$ is the indicator function, and $dim(\theta_l)$ is the number of elements in $\theta_l$. \new{Our goal is to find $\kappa_l$ that guarantees the number of parameter updates on a quantized layer $\Delta \theta^q_{l,k}$ exceeds a certain ratio $T$.}

\newdel{Finding the exact solution of $\kappa_l$ for each layer can be time-consuming because it requires analyzing the distributions of $g_{l,k}$.}
To quickly find an approximate solution, we applied a simple two-step heuristic. 
Firstly, starting from 1.0, $\kappa_l$ is doubled until $\Delta \theta^q_{l,k} > T$. 
Then, to satisfy Eq. \ref{eq:GI_objective}, $\kappa_l$ is adjusted between $\nicefrac{\kappa}{2}$ and $\kappa$ by binary search.
For the sake of computation efficiency, the total number of search steps is limited to five. 
In addition, to assure early phase stability of the training, we added a warm-up phase for the \lrabb method.  
In the warm-up phase, the maximum \newdel{adjustment }of $\kappa$ is limited to 128\newdel{ times larger than the original magnitude} for more accurate solutions.
When the generator requires a separate warm-up, the \lrabb warm-up phase starts after the generator warm-up ends. \JL{KH:chk this entire paragraph}
Similar to learning rate exponential decay scheduling, we apply the exponential decay to $\rho$.
We discuss the sensitivity to this in \cref{sec:sensitivity}.

\section{Experimental Results}
\label{sec:exp}
\subsection{Experimental Environments}
We evaluate \aname on three datasets, CIFAR-10, CIFAR-100~\cite{cifar}, and ImageNet (ILSVRC2012~\cite{imagenet}).
CIFAR-10 and CIFAR-100 contain 10 and 100 classes of images, respectively, and represent small-scale datasets.
ImageNet has 1000 classes of images with 1.2M training samples and 50K validation samples, which represent large-scale dataset\newdel{ in our experiments}.

For CIFAR-10\newdel{ and CIFAR-}\new{/}100, we use commonly used ResNet-20~\cite{resnet} model.
For ImageNet, we use ResNet-18 and ResNet-50 to represent popular medium- and large-sized models, and MobileNetV2~\cite{mobilenetv2} to represent a lightweight model.
All \new{pre-trained }models\newdel{ and full-precision weight parameters} are from pytorchcv library~\cite{pytorchcv}.
For more results on various models, please refer to the Appendix.

For baselines, we use the official code provided by the authors of ZeroQ~\cite{zeroq}, GDFQ~\cite{gdfq}, ARC~\cite{autorecon} and Qimera~\cite{qimera} with the identical settings.
\aname is implemented using PyTorch~\cite{pytorch} version 1.10.0\newdel{, which is installed in Ubuntu 20.04.3 LTS with CUDA version 11.1}.
\new{
All experiments are conducted using NVIDIA RTX3090 and A6000 GPUs.
}

The generator is trained with the loss function \cref{eq:gdfq_G} with $\alpha=0.5$ using Adam optimizer~\cite{adam} with learning rate of 0.001.
For training the quantized student model, SGD with Nesterov~\cite{nesterov} was used with momentum 0.9\JL{did we remove w decay?}.
For \aname, the hyperparameter $\rho$ was set to 0.001 and 0.0001 for CIFAR and ImageNet respectively, which both were decayed by 0.1 every 100 epochs.
\newdel{Experiments on CIFAR-10 and CIFAR-100 were run for 400 epochs with 200 batches on learning rate $\eta=$1e-4, and experiments on ImageNet were run for 400 epochs with 16 batches on learning rate $\eta=$1e-4.}
\new{Experiments on CIFAR and ImageNet were run for 400 epochs on learning rate $\eta=$1e-4, with batchsize of 200 and 16, respectively.}

\label{sec:environment}
\subsection{Performance Comparison}
\label{sec:perf}
\aname can be applied to most generative zero-shot quantization methods.
In this section, we apply our method to three: GDFQ~\cite{gdfq}, the first method to suggest such approach, ARC\cite{autorecon}, which improves the generator, and Qimera~\cite{qimera}, the SOTA technique in the same family. 
We also include ZeroQ~\cite{zeroq} for comparison.
\newdel{For the baselines, we use the official code repository provided by the authors.}
We report top-1 accuracies.

Overall, \aname achieves significant performance improvements in most settings tested, whether implemented on top of GDFQ, ARC or Qimera.
Notably large improvements have been observed on ImageNet datasets, especially in 4w4a settings, because there still exists a large gap towards the full-precision (32bit) model. 
The largest gain was found for 4w4a ResNet-50 on top of GDFQ, with the gain of 12.12\%p that seems to mainly come from the huge gap (25.61\%p) GDFQ originally had between the full-precision model.
\new{For the results of lower-bit settings, refer to Appendix.}

An interesting trend is that for the other two methods with better generators (Qimera, ARC),
the performance gain on 4w4a setting is larger for smaller models (ResNet-50\textrightarrow ResNet-18\textrightarrow MobileNetV2).
The improvements are ($+$1.38\%p, $+$2.99\%p, $+$5.19\%p) for Qimera and ($+$3.90\%p, $+$4.41\%p, $+$6.34\%p) for ARC in \newdel{the order of descending}\new{a descending order of} model size.
This indirectly supports our claim that quantized networks with smaller capacity have difficulties optimizing for multiple loss terms, and \aname can alleviate such effect.

In addition, the performance of ARC+\aname is better than Qimera+\aname for all ImageNet settings except one, even though \newdel{ baseline Qimera outperforms baseline ARC.}\new{ Qimera outperforms ARC in their default settings.}
We find the reason from generator model size of ARC.
While Qimera uses the exact same generator from GDFQ, ARC uses a larger generator model found by neural architecture search. 
The result demonstrate that \aname is making better use of the potential of the generator network.
%
Small performance degradations were observed for CIFAR-10 on Qimera by 0.03\%p.
Since CIFAR-10 is a small dataset and the performance is already close to the fp32 model, we believe this is because there is not much room left to improve.

\begin{table}[b]

    \centering

    \resizebox{\columnwidth}{!}
    {
    \begin{tabular}{cccccc}
    \toprule
    Dataset & \multicolumn{1}{c}{CIFAR-10} & \multicolumn{1}{c}{CIFAR-100} & \multicolumn{3}{c}{ImageNet} \\
    \cmidrule(lr){2-2}\cmidrule(lr){3-3}\cmidrule(lr){4-6}
    Model & RN-20 & RN-20 & RN-18 & RN-50 & MB-V2  \\
    \midrule

Baseline (GDFQ)	&	90.25	&	63.39	&	60.60	&	52.12	&	59.43	\\
KL-only	&	90.06	&	58.93	&	58.49	&	42.64	&	47.03	\\
KL-only (high lr)	&	\textbf{92.20}	&	62.20	&	65.34	&	61.68	&	64.70	\\
Baseline + GI	&	89.32	&	59.05	&	55.01	&	44.09	&	43.57	\\
CE-only + GI	&	90.89	&	51.57	&	52.72	&	27.86	&	33.88	\\
\midrule
AIT (KL-only + GI)	&	91.23	&	\textbf{65.80}	&	\textbf{65.51}	&	\textbf{64.24}	&	\textbf{65.39}	\\

      \bottomrule
    \end{tabular} 
    } 
    \vspace{-2mm}
    \caption{Ablation Study.}
    \vspace{-3mm}
    \label{tab:ablation}
\end{table}

\subsection{Ablation Study}
\label{sec:ablation}
\cref{tab:ablation} shows an ablation study performed over GDFQ.
\new{The ResNet family and MobileNet are denoted as `RN' and `MB' respectively.}
`KL-only' drops CE from the original loss function of GDFQ, and let the quantized model optimize only on the KL divergence against the full-precision teacher.
However, this results in a huge degradation in all settings. 
As analyzed in \cref{sec:kl}, this is due to the scarce, unbalanced quantized weight updates.
By applying \lrname, the lost performance is more than recovered and the superior gain over the baseline is obtained (KL-only+\lrabb).

`Baseline+\lrabb' represents the \lrname applied on top of GDFQ without dropping the CE term,
and `CE only+\lrabb' represents the same with KL term dropped from the loss.
Unfortunately, they only result in performance degradation, because the baseline GDFQ with CE+KL loss or CE loss does not suffer from the aforementioned scarce weight update problems. 
Therefore, \lrname only makes detrimental changes to the quantized weights.
\newcommand{\updatewidth}{0.48\columnwidth}
\begin{figure*}
\centering
\begin{subfigure}[t]{0.45\columnwidth}
\includegraphics[width=\textwidth, trim={0 35mm 0 0}, clip=true]{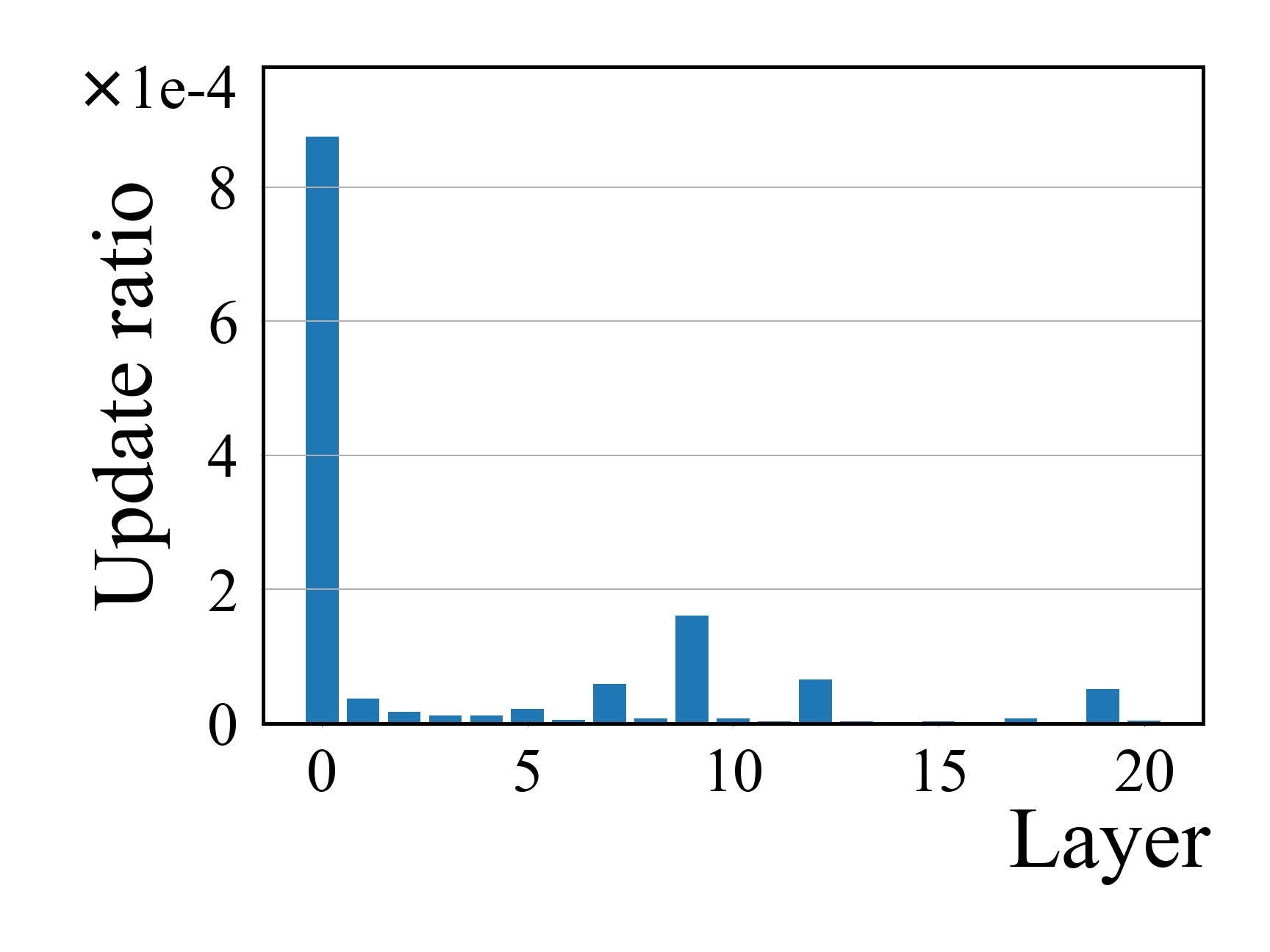}
\caption{}
\label{fig:highlr_60}
\end{subfigure}
\begin{subfigure}[t]{0.45\columnwidth}
\includegraphics[width=\textwidth, trim={0 35mm 0 0},  clip=true]{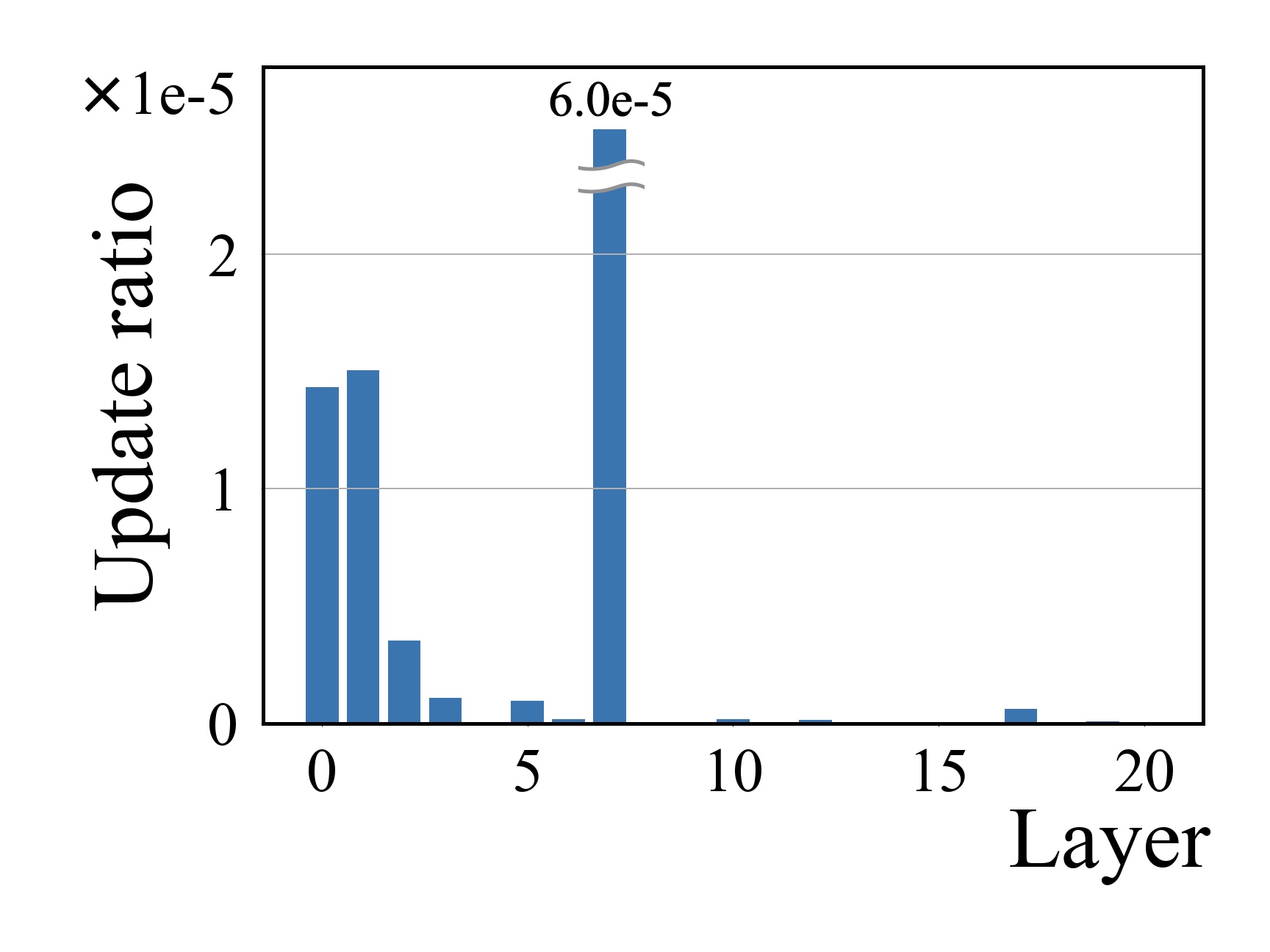}
\caption{}
\label{fig:highlr_350}
\end{subfigure}
\begin{subfigure}[t]{0.45\columnwidth}
\includegraphics[width=\textwidth, trim={0 35mm 0 0},  clip=true]{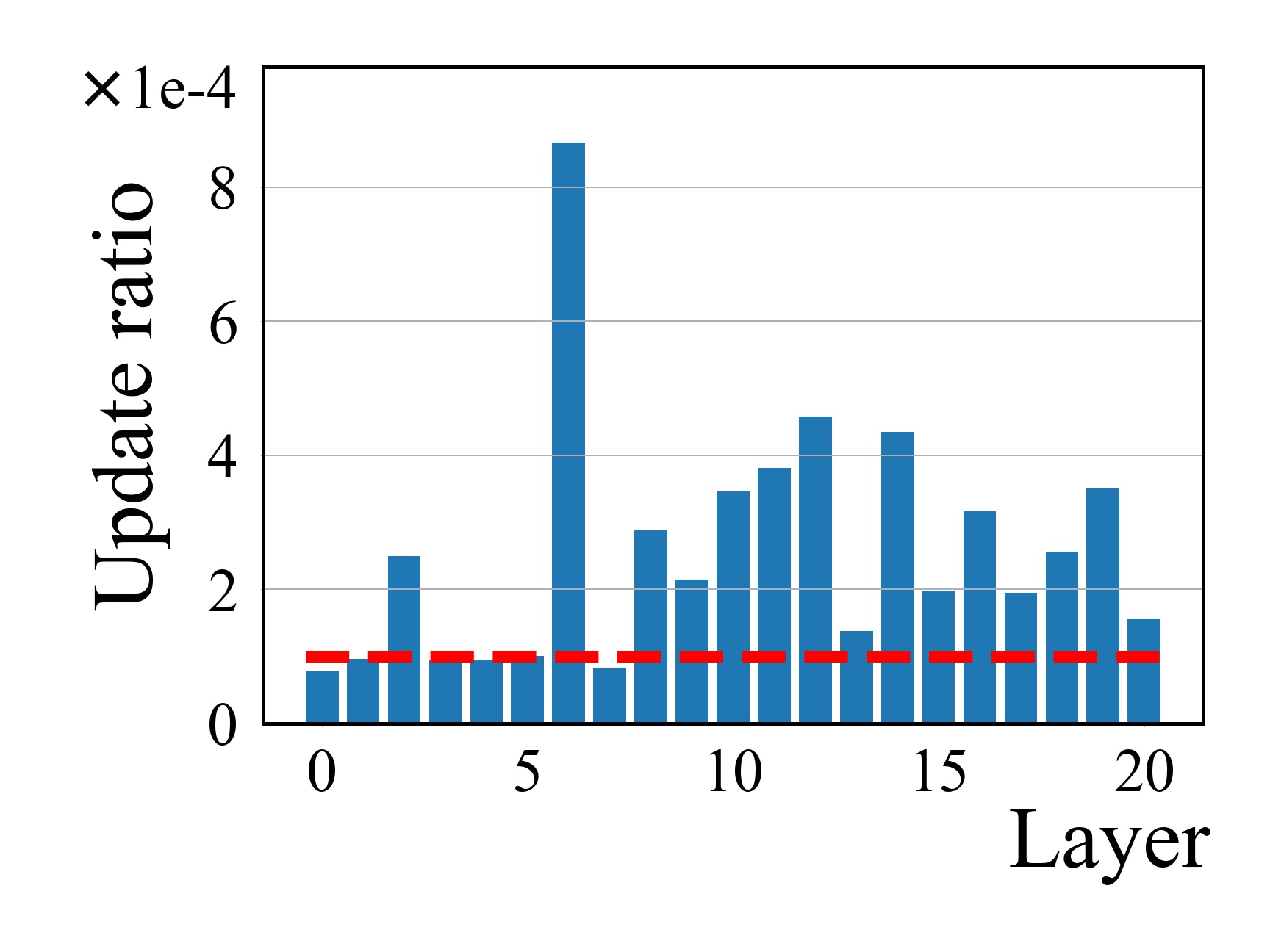}
\caption{}
\label{fig:gi_60}
\end{subfigure}
\begin{subfigure}[t]{0.45\columnwidth}
\includegraphics[width=\textwidth, trim={0 35mm 0 0},  clip=true]{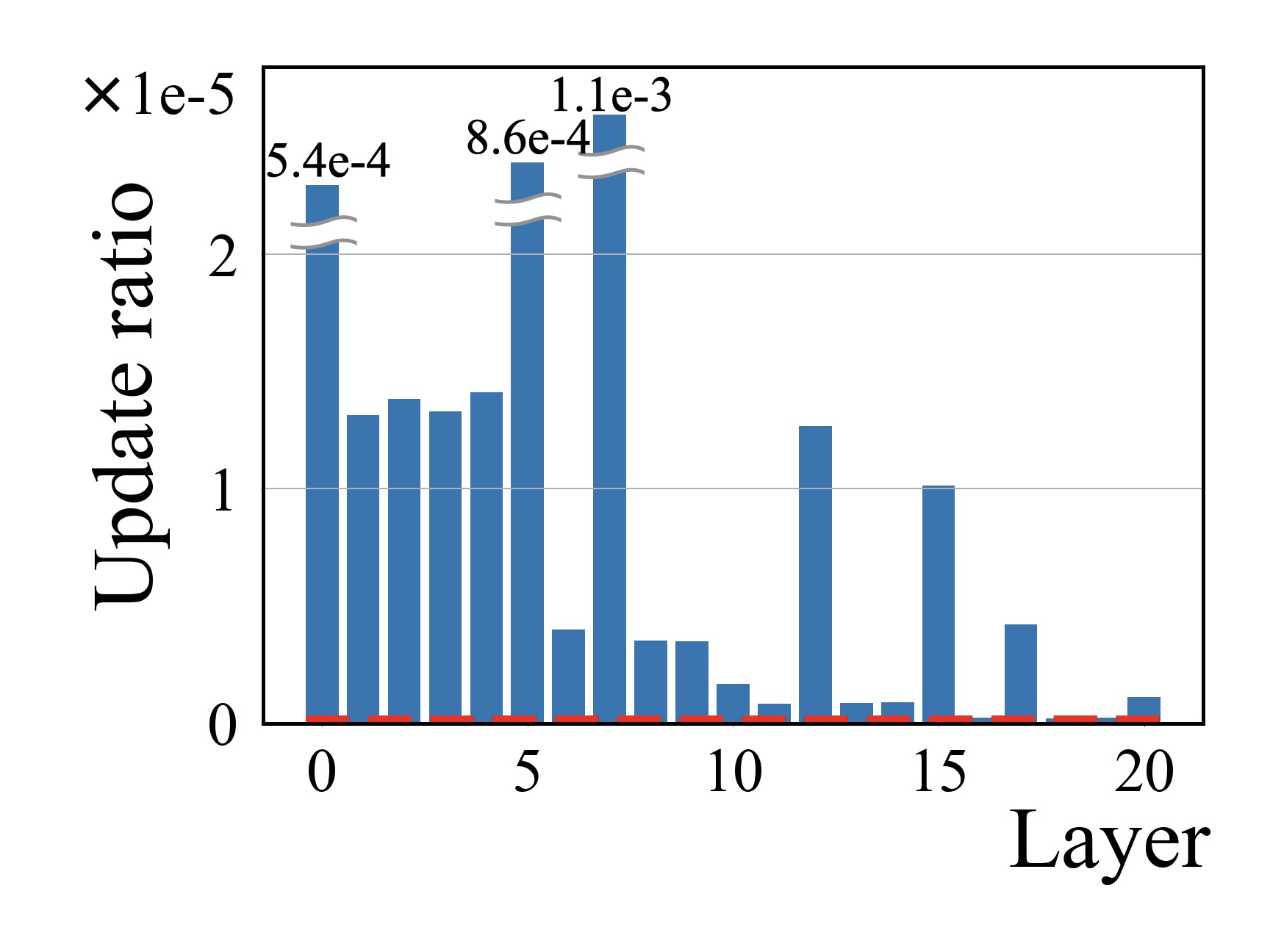}
\caption{}
\label{fig:gi_350}
\end{subfigure}
\vspace{-3mm}
\caption{Distribution of the updated quantized values with $\times$100 learning rate at an earlier epoch (a), later epoch (b) and with \lrname at an earlier epoch (c), later epoch (d).\vspace{-3mm}}
\label{fig:lr_updates}
\end{figure*}

Seeing the effect of `KL-only + \lrabb', one might wonder if the weight update problem can be solved by simply increasing the learning rate.
`KL-only (high lr)' row shows the results of such experiments conducted with $\times$100 learning rate. 
In addition, \cref{fig:lr_updates} shows the distribution of the updates in each layer, in comparison to \cref{fig:count_after}.
\newdel{Increasing the learning rate }\new{`KL-only (high lr)'} achieves a small gain \newdel{.
However, increasing the learning rate}\new{ but} does not entirely solve the problem. 
First, increasing the learning rate incurs too frequent updates in a few layer which was already getting enough updates as shown in \cref{fig:highlr_60,fig:highlr_350}, and many layers still not being updated. 
Moreover, further increasing the learning rate results in divergence of the model.
\cref{fig:gi_60,fig:gi_350} presents the number of updates with \lrname, where $\rho$ is depicted in dotted red lines. \Lrname tunes the gradients to the right level, leading to a better performance.

\begin{figure*}
\centering
\begin{subfigure}[t]{\subfigurewidth}
\centering
\includegraphics[width=\textwidth]{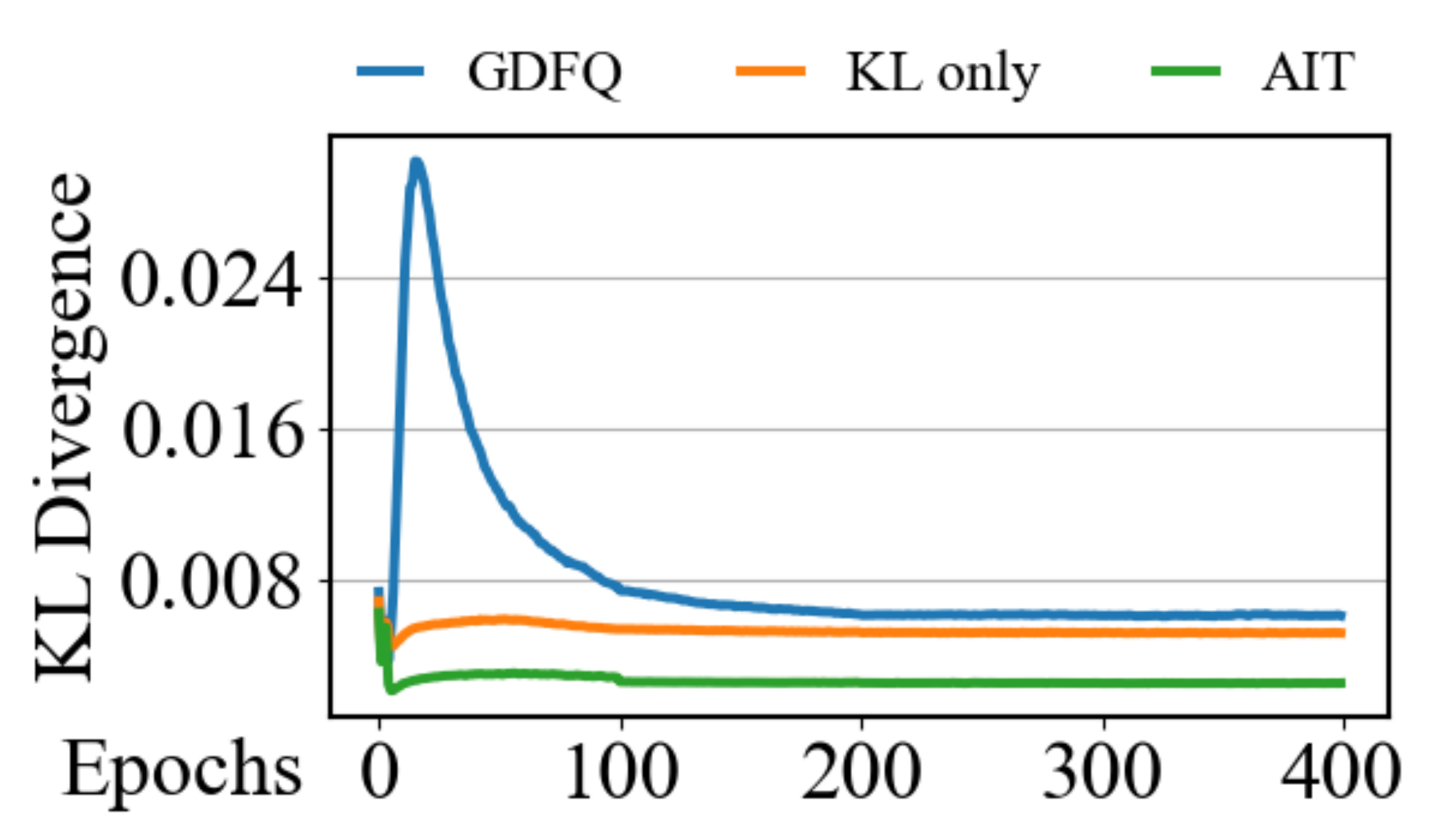}
\caption{}\label{fig:after_kl}
\end{subfigure}
\hspace{3mm}
\begin{subfigure}[t]{\subfigurewidth}
    \includegraphics[width=\columnwidth]{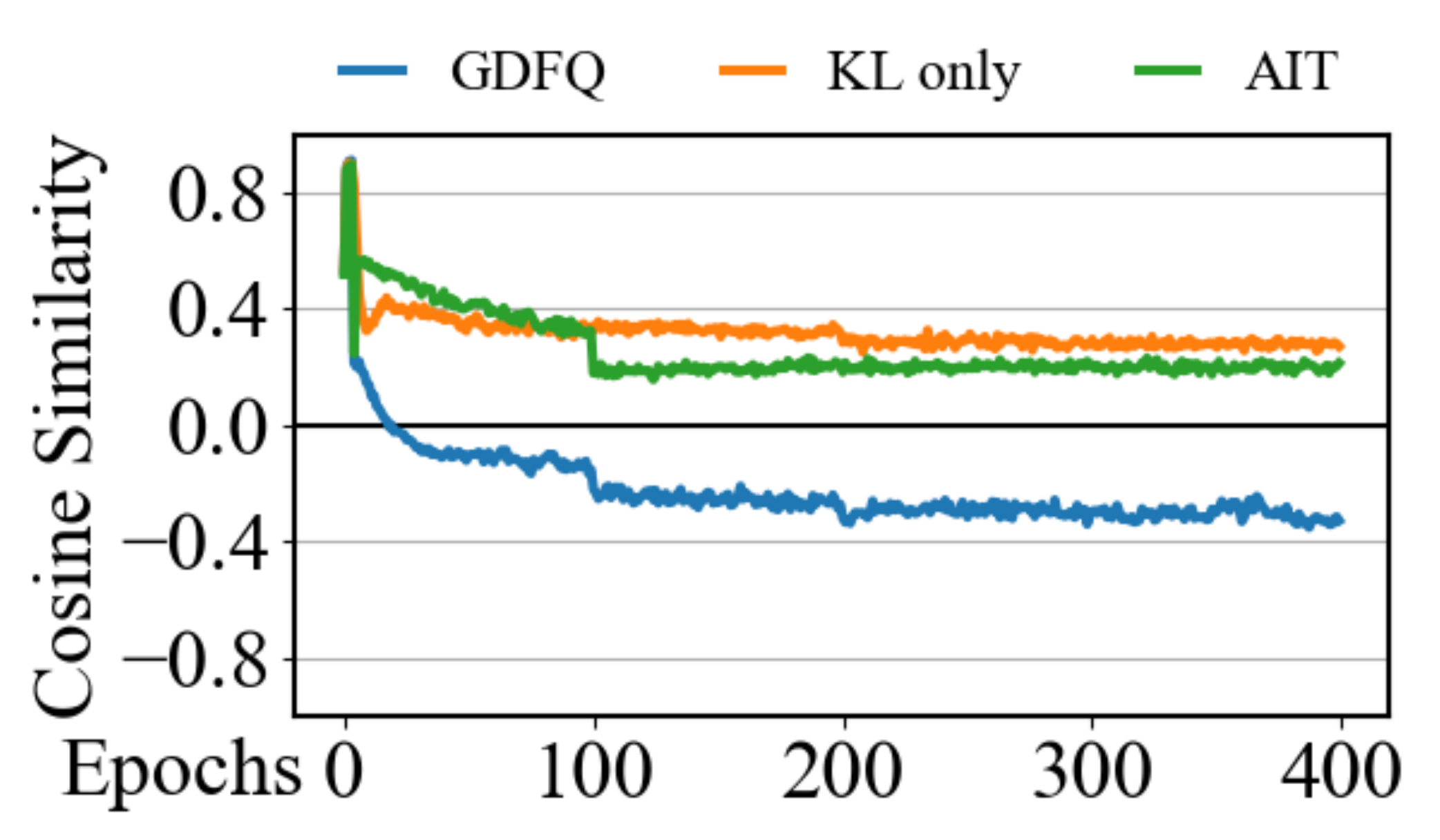}
    \caption{}\label{fig:after_cossim}
\end{subfigure}
\hspace{3mm}
\begin{subfigure}[t]{\subfigurewidth}
\centering
\includegraphics[width=\textwidth]{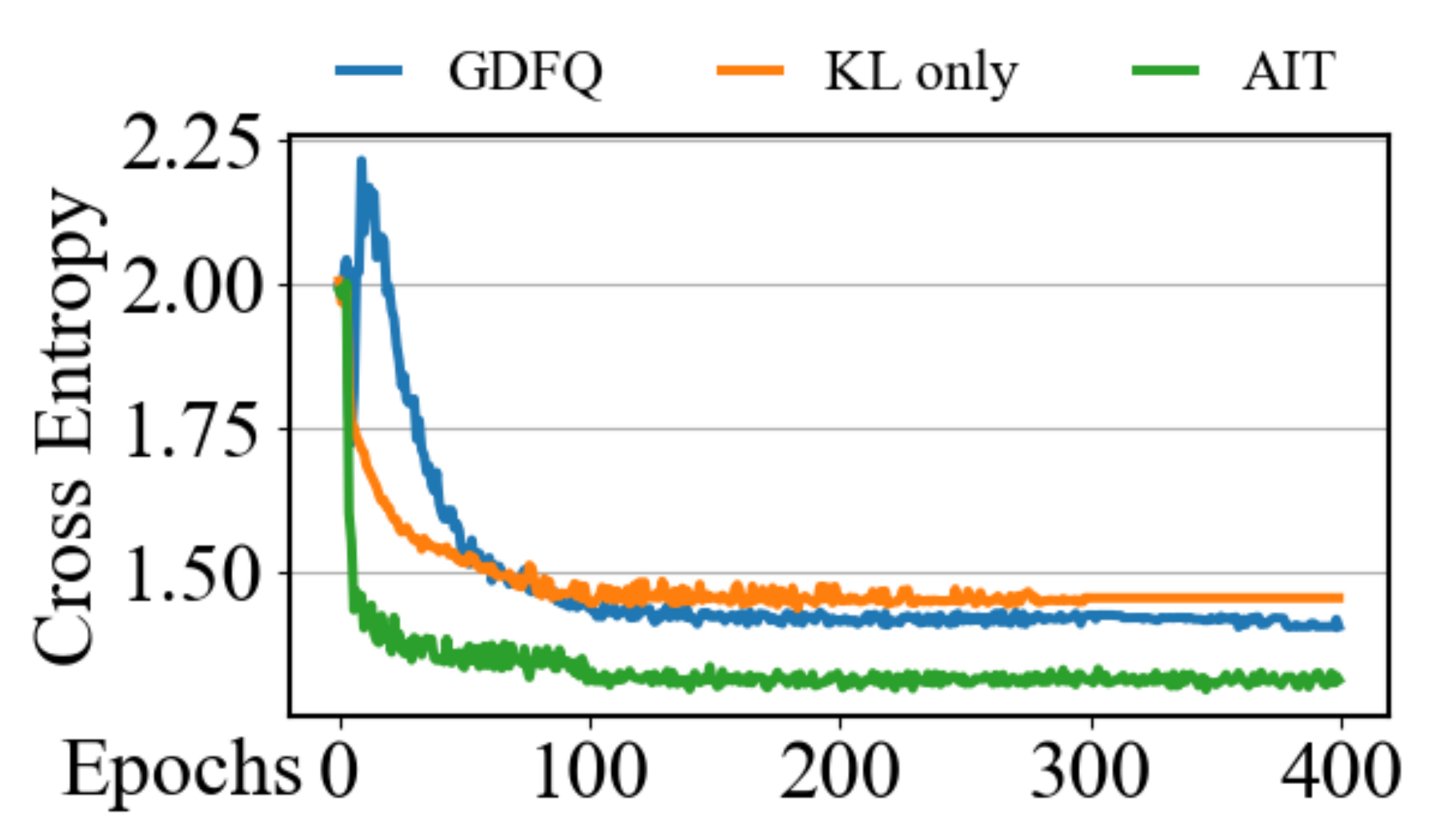}
\caption{}\label{fig:after_ce}
\end{subfigure}
\vspace{-3mm}
\caption{Further experiments. (a) KL divergence. (b) Cosine similarity of KL and CE for GDFQ, KL-only, and \aname.  (c) Cross-entropy against the validation samples. }
\vspace{-2mm}
\label{fig:ait_detail}
\end{figure*}

\newcommand{\samplewidth}{0.22\columnwidth}
\begin{figure}[t]
    \centering
    \subcaptionbox{\label{fig:cifar_samples}}{ 
        \centering
        \includegraphics[height=\samplewidth]{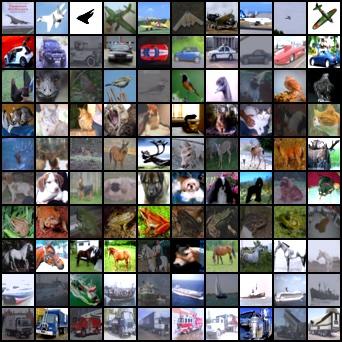} 
    }
     \subcaptionbox{\label{fig:ait_samples}}{ 
         \centering
         \includegraphics[height=\samplewidth]{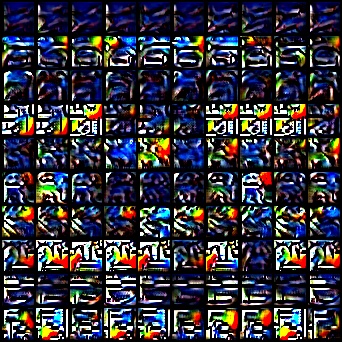} 
     }
    \subcaptionbox{\label{fig:imgnet_samples}}{ 
        \centering
        \includegraphics[height=\samplewidth]{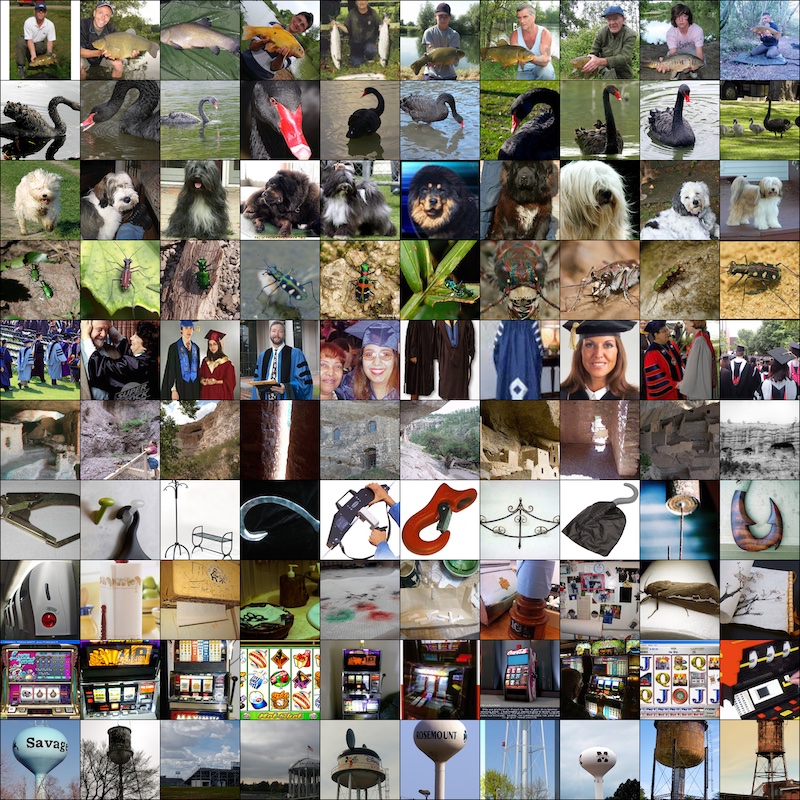} 
    }
     \subcaptionbox{\label{fig:ait_imgnet}}{ 
         \centering
         \includegraphics[height=\samplewidth]{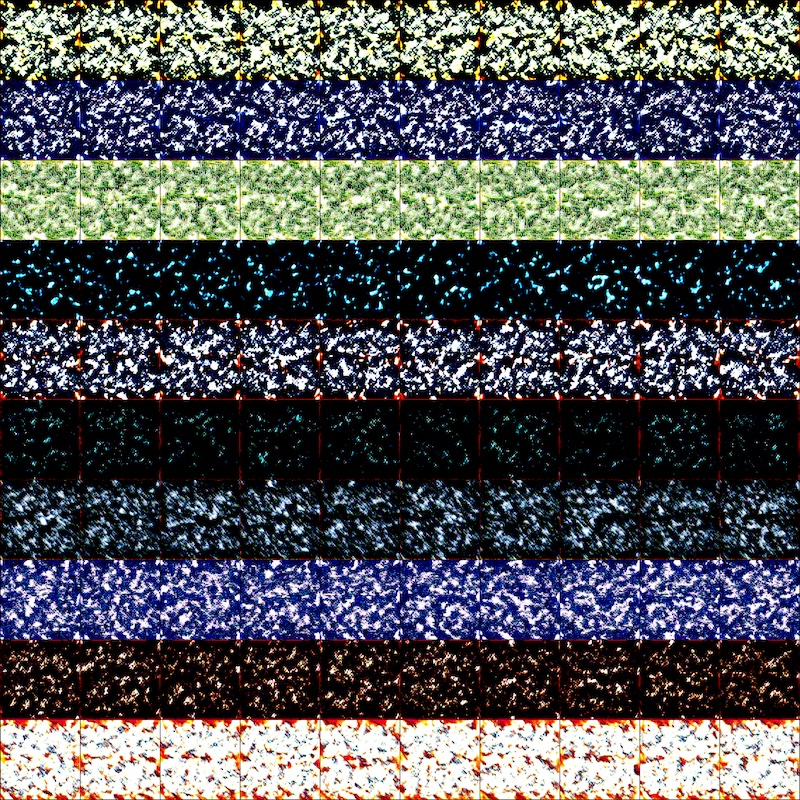} 
     }
    \caption{Comparison of samples. (a) Real CIFAR-10 samples (b) synthetic CIFAR-10 samples (c) Real ImageNet samples (d) synthetic ImageNet samples. Each row represents one of the 10 classes for CIFAR-10, and 10 randomly chosen classes of ImageNet.}
    \vspace{-4mm}
    \label{fig:samples}
\end{figure}

\new{

\begin{table}[t!]
    \centering

     \resizebox{\columnwidth}{!}
     {
    \begin{tabular}{cccccc}
    \toprule
    Dataset & \multicolumn{1}{c}{CIFAR-10} &\multicolumn{1}{c}{CIFAR-100} & \multicolumn{3}{c}{ImageNet} \\
    \cmidrule(lr){2-2}\cmidrule(lr){3-3}\cmidrule(lr){4-6}
    Model & RN-20 & RN-20 & RN-18 & RN-50 & MB-V2  \\
    \midrule

Baseline-LARS   	&	90.01	&	63.84  &	58.94	&	52.98 &	59.58		\\
\midrule
Baseline-SGD    	&	90.25	&	63.39  &	60.60	&	52.12 &	59.43		\\
Baseline-Adam   	&	91.12	&	57.39  &	40.97	&	30.16 &	26.35		\\
Baseline-RMSProp	&	89.88	&	63.18  	&	51.12 &	40.65 & 31.52		\\
\midrule
GI-SGD (ours)   	&	91.23	&	\textbf{65.80}  &	65.51	&	64.24 &	65.39		\\
GI-Adam (ours)  	&	\textbf{91.33}	&	64.38  &	65.47	&	\textbf{65.67} &	61.33		\\
GI-RMSProp (ours)	&	90.82	&	65.78  &	\textbf{65.73}	&	62.42 &	\textbf{65.41}		\\

      \bottomrule
    \end{tabular} 
     } 
    \vspace{-2mm}
    \caption{GI on Various Optimizers.}
    \vspace{-3mm}
    \label{tab:optimizer}
\end{table}


For a comprehensive comparison, we also tested adaptive optimizers such as Adam~\cite{adam}, RMSProp~\cite{tieleman2012lecture}, and especially LARS~\cite{you2017large}, which adjusts learning rates per layer. 
The results of these optimizers for the quantized model are shown in \cref{tab:optimizer}.
`Baseline-' denotes the existing methods, and `GI-' denotes modified optimizer with GI.
The results show that `Baseline-Adam' and `Baseline-RMSProp' suffer from noticeable accuracy degradation, especially on the large models.
`Baseline-LARS' survived from such trend, it does not make significant differences compared to `Baseline-SGD'.
We further expand our research by applying GI to Adam and RMSProp, written as `GI-Adam' and `GI-RMSProp'. 
The results verify that our method has solid performance on various optimizers and still outperforms existing methods. 

}

\subsection{Sensitivity Analysis}
\label{sec:sensitivity}
\begin{table}[t!]
    \centering

     \resizebox{\columnwidth}{!}
     {
    \begin{tabular}{llccccc}
    \toprule
    \multirow{2}{*}{Dataset} & \multirow{2}{*}{Model} & \multicolumn{5}{c}{$\rho$}\\
    \cmidrule(lr){3-7}
    & & 0.005 & 0.001 & 0.0005 & 0.0001 & 0.00005  \\
    \midrule

CIFAR-100  &ResNet-20   	&	{63.01}	&	{65.80}  &	{65.41}	&	{65.04} &	{65.30}		\\
ImageNet  &ResNet-18   	&	{57.95}	&	{60.11}  &	{64.48}	&	{65.51} &	{65.92}		\\

      \bottomrule
    \end{tabular} 
     } 
    \vspace{-2mm}
    \caption{Sensitivity Analysis.}
    \vspace{-4mm}
    \label{tab:sense}
\end{table}
\begin{table}[b]
    \centering

    \resizebox{0.75\columnwidth}{!}
    {
    \begin{tabular}{ccccc}
    \toprule
    \multirow{2}{*}{$\eta$} & \multicolumn{2}{c}{Cifar-100 (RN-20)} & \multicolumn{2}{c}{ImageNet (RN-18)} \\
    \cmidrule(lr){2-3}\cmidrule(lr){4-5}
     & GDFQ & AIT & GDFQ & AIT  \\
    \midrule

1e-2	&	49.81	&	66.96   &  40.78	&	65.69	\\
					
1e-3	&	58.10	&	66.21   &  40.90	&	65.57	\\
					
1e-4	&	63.39	&	65.80   &  53.28	&	65.51	\\
					
1e-5	&	61.08	&	65.92   &  59.32	&	65.70	\\
					
1e-6	&	59.47	&	65.73   &  60.60	&	65.23	\\
      \bottomrule
    \end{tabular} 
    }
    \vspace{-2mm}
    \caption{Learning Rate Sensitivity Analysis.}
    \vspace{-3mm}
    \label{tab:lr}
\end{table}
The $\rho$ value controls the portion of quantized weights guaranteed to get updates in each layer.
In \cref{tab:sense}, a sensitivity study of $\rho$ on top of GDFQ+AIT has been performed.\newdel{ on a range of $\rho$ values.} 
\newdel{As displayed in \cref{tab:sense},}\new{The results show that} \aname is not very sensitive to $\rho$, although there is some effect.
\new{Refer to the Appendix for more results.}

\new{
Tab 5. shows the learning rate sensitivity of our method with comparison to the GDFQ. 
The results show that the GI method is robust to the learning rate changes on both datasets while steadily outperforms the baseline method.
}

\subsection{Further Analysis}
In this section, we present more details of \aname.
\cref{fig:after_kl} shows the KL divergence over the training in the baseline GDFQ, `KL-only' and \aname.
\aname is able to reach a lower KL distance. 
This supports our analysis from \cref{sec:kl} that there is still room for KL to be optimized further.

Another observation can be found from \cref{fig:after_cossim}, 
where we have measured the angle between the CE and KL.
For `KL-only' and \aname, CE is calculated only for measurements, and did not affect the SGD updates.
It is now shown that `KL-only' and \aname, both have positive cosine similarities between the losses. 
This represents that, although they are difficult to optimize concurrently in the beginning, as they get near the global minima, they become to share the same direction for optimization. 
\aname sacrifices a small amount of directional alignment, but outperforms `KL-only' by guaranteeing quantized weight updates with \lrname.
Lastly, \cref{fig:after_ce} shows that \aname achieves lower CE in validation against the hard label, even though it has not seen the real data, and not optimized for CE.

\JL{maybe talk a bit more about update distributions}

\section{Discussion}
\textbf{Removing Cross-Entropy} against the hard label from the loss function as done in \aname \newdel{removes the cross-entropy against the hard label from the loss function.}
\newdel{At a glance, this} could be a penalty, because some methods~\cite{qimera} \newdel{seem to }rely on the sample labels.
However, as we demonstrated in \cref{sec:exp}, \aname applied to Qimera was able to obtain significantly better performance despite the exclusion of the mixed labels.
In addition, our method does not depend on \newdel{whether the predicted label is ground truth or not}\new{per-image hard label}, so it can be widely used for segmentation ~\cite{ronneberger2015u, long2015fully} or object detection~\cite{redmon2016you, ren2015faster}.

\textbf{Privacy Leak} is one societal concern \newdel{One societal concern} of zero-shot quantization \newdel{is privacy leak} because the generator creates synthetic samples that \newdel{closely }follow the distribution of real data.
As several input reconstruction techniques point out~\cite{dosovitskiy2016inverting, mahendran2015understanding}, \newdel{it could be a privacy leak if} \new{it could be that} the synthetic samples can reconstruct the private training data.
However, to the extent of our observation, there is no sign that \aname reconstructs the real data 
as shown in \cref{fig:samples}. 
Training method for the generator in \aname is no different from its baselines, since it does not alter the generator loss in \cref{eq:gdfq_G}, thus does not contribute any further to the privacy leak.

\JL{other regularizer such as label smoothing?}

\section{Conclusion}
In this work, we analyzed the SOTA family of solutions for zero-shot quantization.
Through a series of experiments and analyses, we found that current solutions can be improved through pursuing a flatter minima and guaranteeing weight updates during fine-tuning.
We achieve the goal by bringing the quantized model closer to the full-precision model in terms of the KL divergence and designing \aname --- before our method, people habitually have used a combination of CE and KL as the main loss of the zero-shot quantization.
Experimental results show that \aname is effective and can be easily applied to existing algorithms. 

\section*{Acknowledgements}
\footnotesize{
This work has been supported by the National Research Foundation of Korea (NRF) grant funded by the Korea government (MSIT) (2022R1C1C1008131,  
2022R1C1C1011307),  
Year 2022 Copyright Technology R\&D Program by Ministry of Culture, Sports and Tourism and Korea Creative Content Agency(Project Name: Development of artificial intelligence-based copyright infringement suspicious element detection and alternative material content recommendation technology for educational content, Project Number: CR202104003, Contribution Rate: 30\%), 
and 
Institute of Information \& communications Technology Planning \& Evaluation
(IITP) grant funded by the Korea government (MSIT) 
(2020-0-01361,
Artificial Intelligence Graduate School Program (Yonsei University)
).
}

{\small
\bibliographystyle{ieee_fullname}
\bibliography{egbib}
}

\clearpage
\renewcommand{\thesection}{\Alph{section}}
\setcounter{section}{0}

\section{Code}

As a part of the supplementary materials, the code used to conduct experiments is attached as a separate zip archive. 
The zip archive contains the implementation of the \aname method on multiple zero-shot quantization backbones: GDFQ~\citeApp{gdfq_app}, ARC~\citeApp{autorecon_app}, and Qimera~\citeApp{qimera_app}. 
For reproducibility, the experiment environment setting and training scripts are included for all backbones. 
The code is under the terms of the GNU General Public License v3.0.

\new{
\section{Lower Bit-width Experiments}

\begin{table}[b!]
\renewcommand{\arraystretch}{0.7}
    \centering
    
    \vspace{-5mm}
     \resizebox{\columnwidth}{!}
     {
    \begin{tabular}{ccccccc}
    \toprule
    \multirow{3}{*}{Dataset} & \multirow{3}{*}{Model} & \multirow{3}{*}{Bits}   & \multicolumn{2}{c}{GDFQ} & \multicolumn{2}{c}{ARC} \\
    \cmidrule(lr){4-5}\cmidrule(lr){6-7}
    &&&    Baseline & AIT & Baseline& AIT \\
    \midrule
    \multirow{8}{*}{ImageNet} & \multirow{2}{*}{RN-18}   & 3w3a &20.69&{36.34}&1.00&36.70 \\
                              &                        & 3w4a & 39.73 & 53.55 & 2.54 & {56.77} \\
                              \cmidrule(lr){2-7}
                              & \multirow{2}{*}{RN-50}   & 3w3a & 0.21 & 1.31 & 0.20 & {3.98} \\
                              &                        & 3w4a & 26.85 & 37.50 & 1.37 & {49.34} \\
                              \cmidrule(lr){2-7}
                              & \multirow{2}{*}{MB-V2}   & 3w3a & 5.50 & 13.83 & 0.20 & {30.35} \\
                              &                        & 3w4a & 26.87 & 37.77 & 0.22 & {47.41} \\
                              \midrule
    \multirow{4}{*}{CIFAR-100} & \multirow{4}{*}{RN-20} & 2w2a & 1.41 & {2.09} & 1.35 & 1.55 \\
                              &                        & 2w3a & 1.04 & 1.13 & {1.25} & 1.14 \\
                              &                        & 3w3a & {49.62} & 48.64 & 28.54 & 34.39 \\
                              &                        & 3w4a & 59.70 & {61.37} & 50.47 & 58.65 \\
                              \midrule
    \multirow{4}{*}{CIFAR-10} & \multirow{4}{*}{RN-20} & 2w2a & {16.48} & 15.57 & 16.18 & 13.47 \\
                              &                        & 2w3a & 37.64 & {40.98} & 20.87 & 20.42 \\
                              &                        & 3w3a & {80.70} & 80.49 & 52.99 & 51.78 \\
                              &                        & 3w4a & 90.02 & {90.20} & 82.10 & 82.98 \\
      \bottomrule
    \end{tabular} 
     } 
    \vspace{-2mm}
    \caption{Low Bit-width Experiments Results}
    \vspace{-4mm}
    \label{tab:lowbit_app}
\end{table}

Further experiments on GDFQ and ARC were conducted in lower-bit settings. 
The experiment results are shown in \cref{tab:lowbit_app}
Following the main paper, the ResNet family and MobileNet are denoted as `RN' and `MB', respectively. 
We tested 3w3a and 3w4a quantization settings for the ImageNet experiments and further down to 2w2a and 2w3a for Cifar-10/100, which we found to be the lowest bits GDFQ and AIT converge. 
}

\section{Experiments on Additional Network Models}

\begin{table}[t]

    \centering
    
    \renewcommand{\arraystretch}{0.9}
    \resizebox{0.45\textwidth}{!}
    {
    
    \begin{tabular}{x{1.5cm}|c|c|c|cc|cc|cc}
    \toprule
    Dataset & \makecell{Model\\(FP32 Acc.)} & Bits & 
    GDFQ & \makecell{GDFQ\\{+\aname}}
    \\
    \midrule
    
	\multirow{6}{*}{ImageNet}	&	InceptionV3	&	4w4a	&	70.57	&	73.34	( 	$+$2.77	)\\	
		&	79.00	&	5w5a	&	77.25	&	77.67	( 	$+$0.42	)	\\	
		&	SqueezeNext	&	4w4a	&	26.21	&	45.37	( 	$+$19.16	)		\\	
		&	69.39	&	5w5a	&	56.07	&	62.76	( 	$+$6.69	)	\\	
		&	ShuffleNet	&	4w4a	&	19.72	&	27.80	( 	$+$8.08	)\\	
		&	65.07	&	5w5a	&	45.92	&	48.97	( 	$+$3.05	)\\		
	
    \bottomrule

    \end{tabular}
    }

    \vspace{-2mm}
    \caption{Additional experiments on various network models.}
    \vspace{-4mm}
    \label{tab:morenet_app}
\end{table}
We conducted a further evaluation of our method on various networks: InceptionV3 \citeApp{Szegedy_2016_CVPR_app}, SqueezeNext\citeApp{Gholami_2018_CVPR_Workshops_app}, and ShuffleNet\citeApp{Zhang_2018_CVPR_app}. 
The experimental results are shown in Table \ref{tab:morenet_app}. 
Compared with the GDFQ baseline, our method still outperforms by a huge margin on all settings regardless of the quantization bitwidth. 
Furthermore, experimental results show that \aname is especially effective on smaller networks. 
This result again supports our observation in the main body that the limited capacity of a small network hinders the training phase from matching multiple loss terms simultaneously.

\section{Comparison with Label Smoothing}
\emph{Label smoothing} is a regularization technique that replaces one-hot label $y$ into a  smooth label $y'$ by 
\begin{equation}
    y' = (1-c)y + c / K, \\ 
\end{equation}
where $K$ is the number of classes and $c$ is a label smoothing value.
Label smoothing is known to help neural network training to avoid overfitting and increase generalization capability. 
Therefore, one might think that label smoothing can also help flatten the cross-entropy (CE) loss surface by its nature. 
To answer the question, we conducted comparative experiments with various label smoothing parameters. 
The experiments evaluate how the label smoothing affects the performance of GDFQ baseline and CE-only setting, which drops KL divergence from the training loss. 

Table 2 shows the experimental results. 
For CIFAR-10 and CIFAR-100, label smoothing did not improve performance in any settings over the baseline GDFQ, whether with KL divergence or not.
Some improvements were observed from ImageNet dataset, but the improvements were smaller than that of \aname.
This shows that label smoothing helps flatten the loss surface to some degree, its effect was not enough to reach that of \aname.

\new{
\section{Further Analysis on $\rho$ Sensitivity}

\begin{table}[b!]
    \centering

    \resizebox{\columnwidth}{!}
    {
    \begin{tabular}{cccccc}
    \toprule
    \multirow{2}{*}{$\rho$} & \multicolumn{1}{c}{CIFAR-100} & \multicolumn{1}{c}{ImageNet} & \multirow{2}{*}{$\rho$} & \multicolumn{1}{c}{CIFAR-100} & \multicolumn{1}{c}{ImageNet}\\
    \cmidrule(lr){2-2}\cmidrule(lr){3-3} \cmidrule(lr){5-5}\cmidrule(lr){6-6}
     & RN-20 & RN-18  &  
     & RN-20 & RN-18  \\
    \midrule

0.0005	&	65.41$\pm$0.20	&	64.48$\pm$0.28 &	0.00009	&	65.20$\pm$0.29	&	65.84$\pm$0.07\\
0.0004	&	65.55$\pm$0.15	&	65.23$\pm$0.10 &	0.00008	&	65.29$\pm$0.19	&	65.66$\pm$0.17\\
0.0003	&	65.44$\pm$0.34	&	65.41$\pm$0.53 &	0.00007	&	65.35$\pm$0.18	&	65.65$\pm$0.05\\
0.0002	&	65.21$\pm$0.27	&	65.85$\pm$0.07 &	0.00006	&	65.06$\pm$0.23	&	65.52$\pm$0.16\\
0.0001	&	65.04$\pm$0.13	&	65.51$\pm$0.09 &	0.00005	&	65.30$\pm$0.10	&	65.92$\pm$0.42\\
      \bottomrule
    \end{tabular} 
    }
    \vspace{-2mm}
    \caption{Sensitivity Analysis on $\rho$.}
    \vspace{-4mm}
    \label{tab:rho_app}
\end{table}

We deepen the sensitivity analysis with finer levels of $\rho$ values. 
The experiments are conducted five times per setting to demonstrate performance stability regarding $\rho$ values. 
The results in \cref{tab:rho_app} show that our method can achieve a stable accuracy level without hand-crafted hyperparameter tuning.
}

\begin{table}[b]
    \centering
    \label{tab:noise}    
    \resizebox{\columnwidth}{!}
    {
    \begin{tabular}{cccccccc}
    \toprule
    \multirow{2}{*}{Dataset} & \multirow{2}{*}{Model}  &\multirow{2}{*}{Method} & \multicolumn{4}{c}{$c^*$} & \multirow{2}{*}{\aname}  \\
    \cmidrule(lr){4-7} 
                           &                             &                     &  0.00$^\dagger$ & 0.10 & 0.30 & 0.50  & \\ 
    \midrule
\multirow{2}{*}{CIFAR-10}  & \multirow{2}{*}{RN-20}  & Baseline & 90.25 & 89.67 & 88.85 & 88.52 &  \multirow{2}{*}{91.23}\\
                           &                             & CE only & 88.36 & 88.67 & 88.21 & 87.89 & \\
\multirow{2}{*}{CIFAR-100} & \multirow{2}{*}{RN-20}  & Baseline  &63.39 & 60.50 & 59.11 & 58.53 & \multirow{2}{*}{65.80}\\
                           &                             & CE only & 56.76 & 60.10 & 59.13 & 57.81 & \\
\multirow{2}{*}{ImageNet}  & \multirow{2}{*}{RN-18}  & Baseline & 60.60 & 62.41 & 62.57 & 62.25 &\multirow{2}{*}{65.51}\\
                           &                             & CE only & 60.33&62.48& 62.27 & 62.18 & \\
                                     
    \bottomrule
    \multicolumn{8}{r}{$^\dagger$No smoothing *Label smoothing parameter.}

    \end{tabular}
    }
    \vspace{-2mm}
    \caption{Performance of GDFQ with label smoothing in 4w4a setting.}
    \vspace{-4mm}
\end{table}

\section{Gradient Cosine Similarity}
Although the main body of the manuscript offers results for gradient cosine similarity measured on ResNet20 with CIFAR-10 dataset, we have done an extensive amount of experiments to study the distinct gradient directionality spotted in zero-shot quantization task. 
Here we share the results to further support our findings. 

For CIFAR-10 and CIFAR-100 dataset, we used ResNet-20, ResNet-56, ResNeXt-29 32x4d, WRN28-10, and WRN40-8. 
On ImageNet, we evaluated on ResNet-18, ResNet-50, MobileNetV2, and InceptionV3. 
The experiment compares the directionality of loss functions in training these networks under two different settings: zero-shot quantization (ZQ) and knowledge distillation (KD). 
In the knowledge distillation setting, we used the same network for both the student and the teacher (self-distillation) for fair comparison against the Zero-shot quantization setting.

\begin{figure*}[t]
\centering
\begin{subfigure}[t]{0.37\textwidth}
\centering
\includegraphics[width=0.9\textwidth]{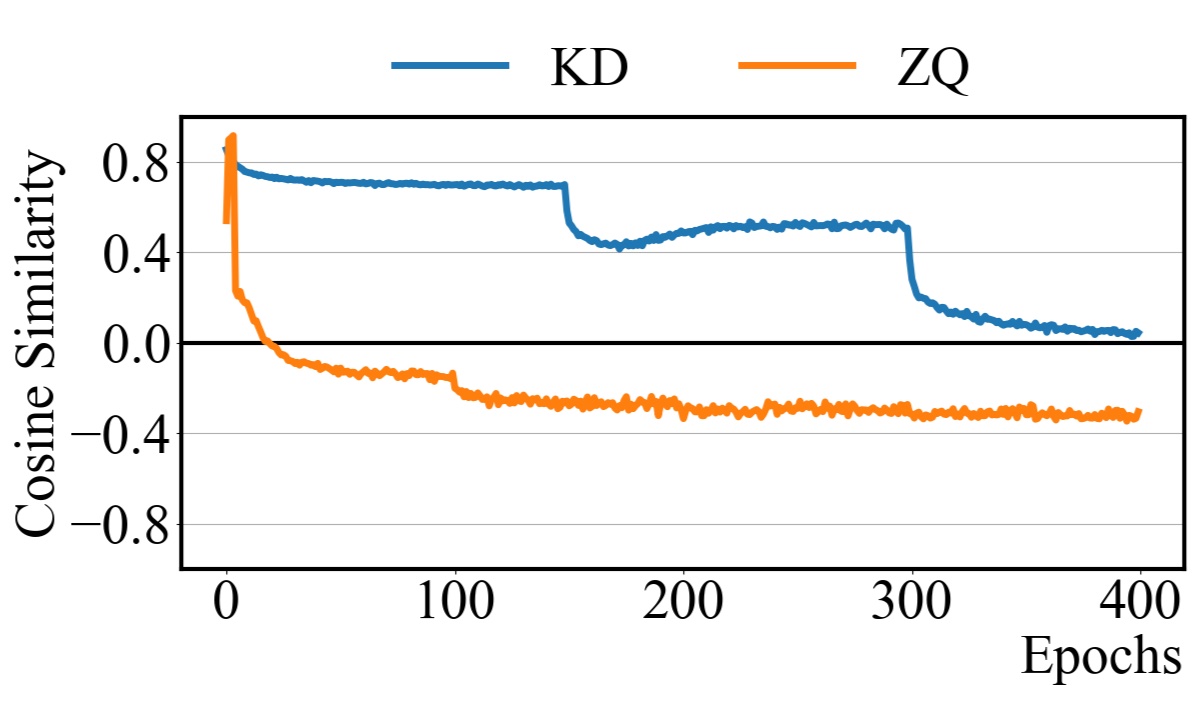}
\includegraphics[width=0.45\textwidth]{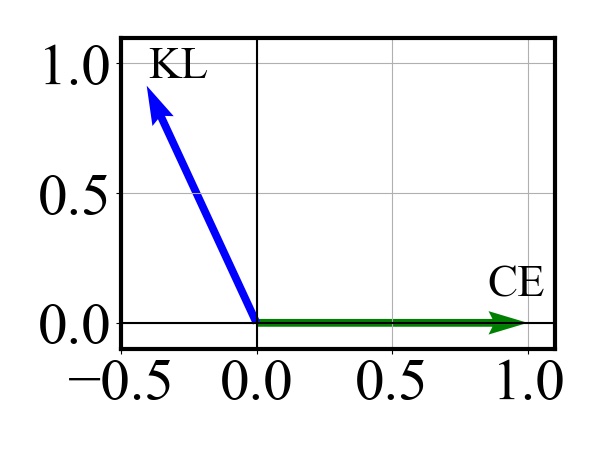}
\includegraphics[width=0.45\textwidth]{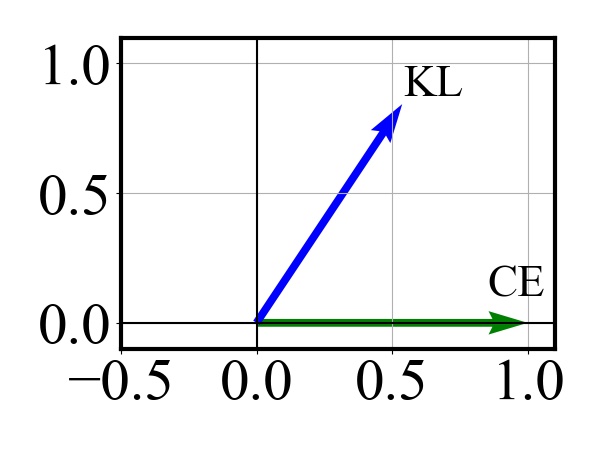}
\caption{CIFAR-10 ResNet-20}
\end{subfigure}
\hspace{10mm}
\begin{subfigure}[t]{0.37\textwidth}
\centering
\includegraphics[width=0.9\textwidth]{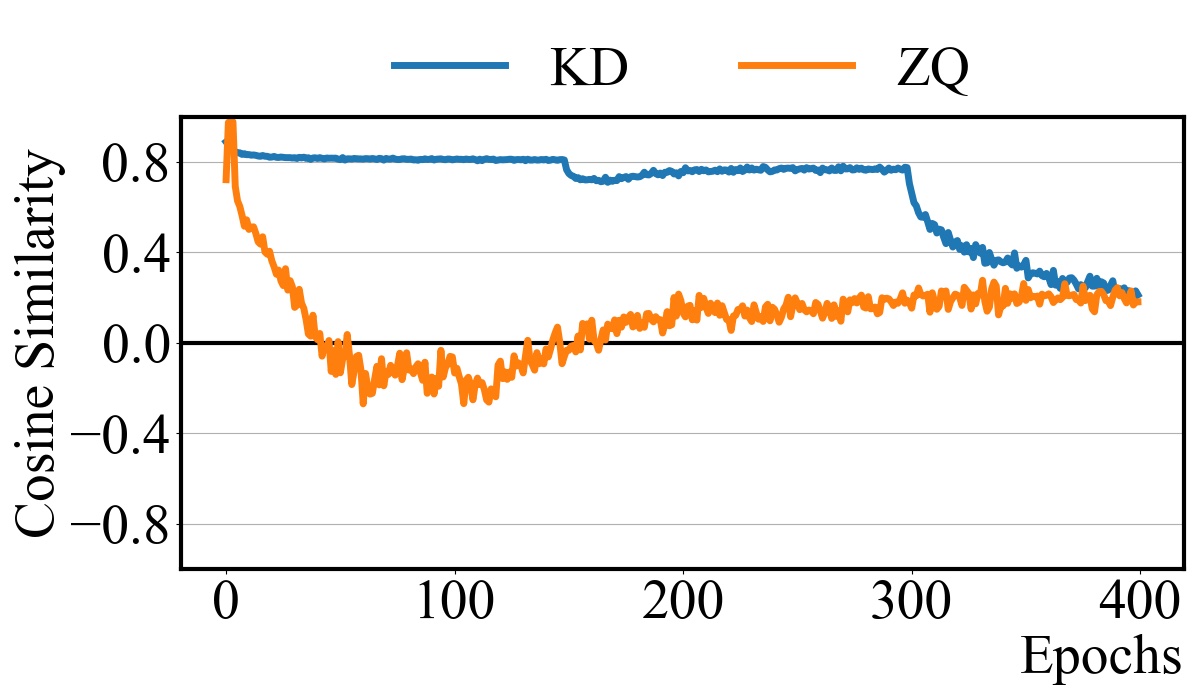}
\includegraphics[width=0.45\textwidth]{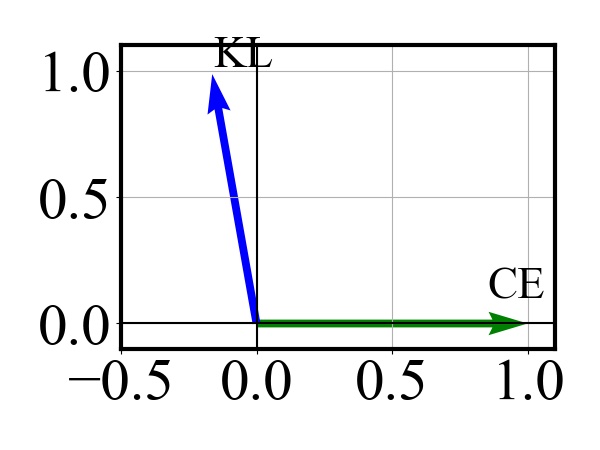}
\includegraphics[width=0.45\textwidth]{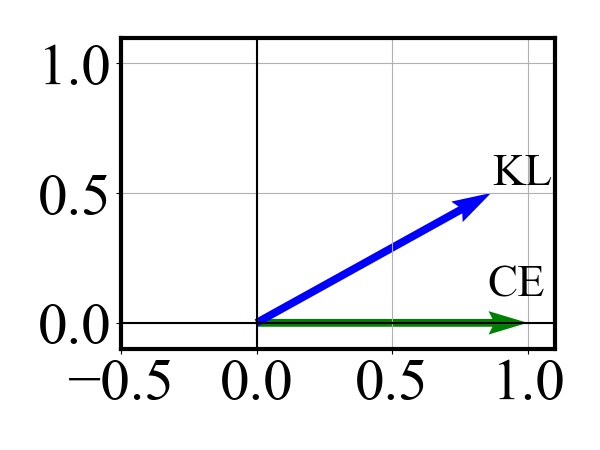}
\caption{CIFAR-10 ResNet-56}
\end{subfigure}
\begin{subfigure}[t]{0.37\textwidth}
\centering
\includegraphics[width=0.9\textwidth]{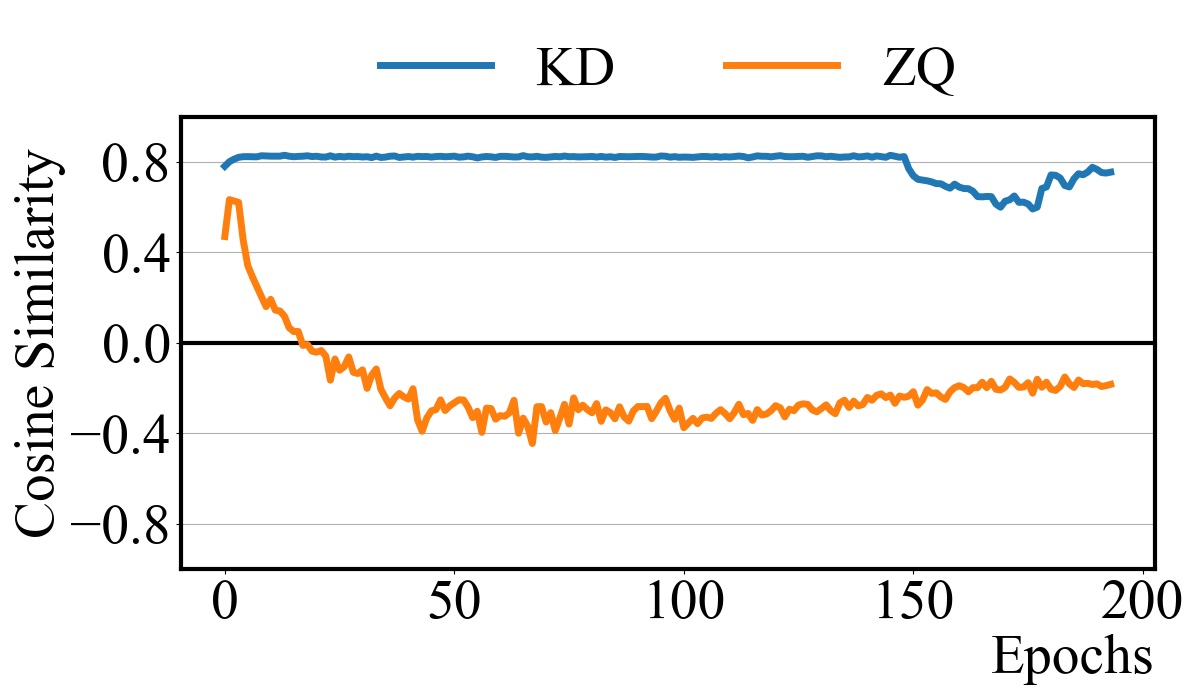}
\includegraphics[width=0.45\textwidth]{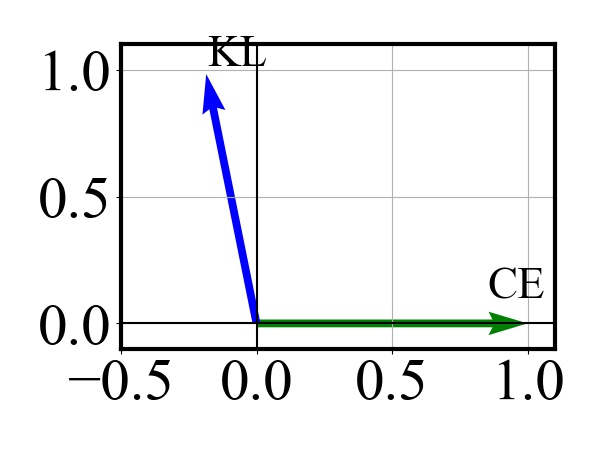}
\includegraphics[width=0.45\textwidth]{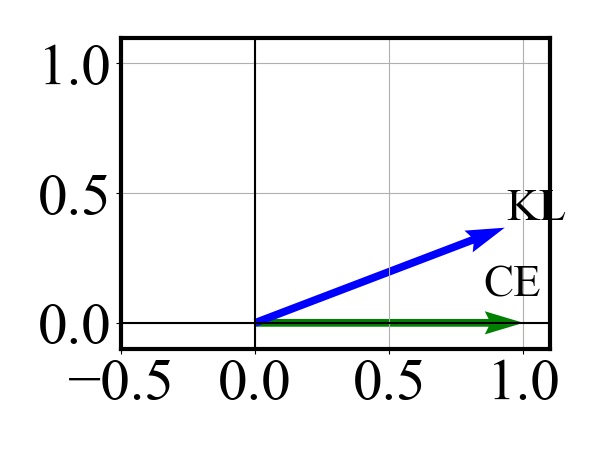}
\caption{CIFAR-10 WRN28-10}
\end{subfigure}
\hspace{10mm}
\begin{subfigure}[t]{0.37\textwidth}
\centering
\includegraphics[width=0.9\textwidth]{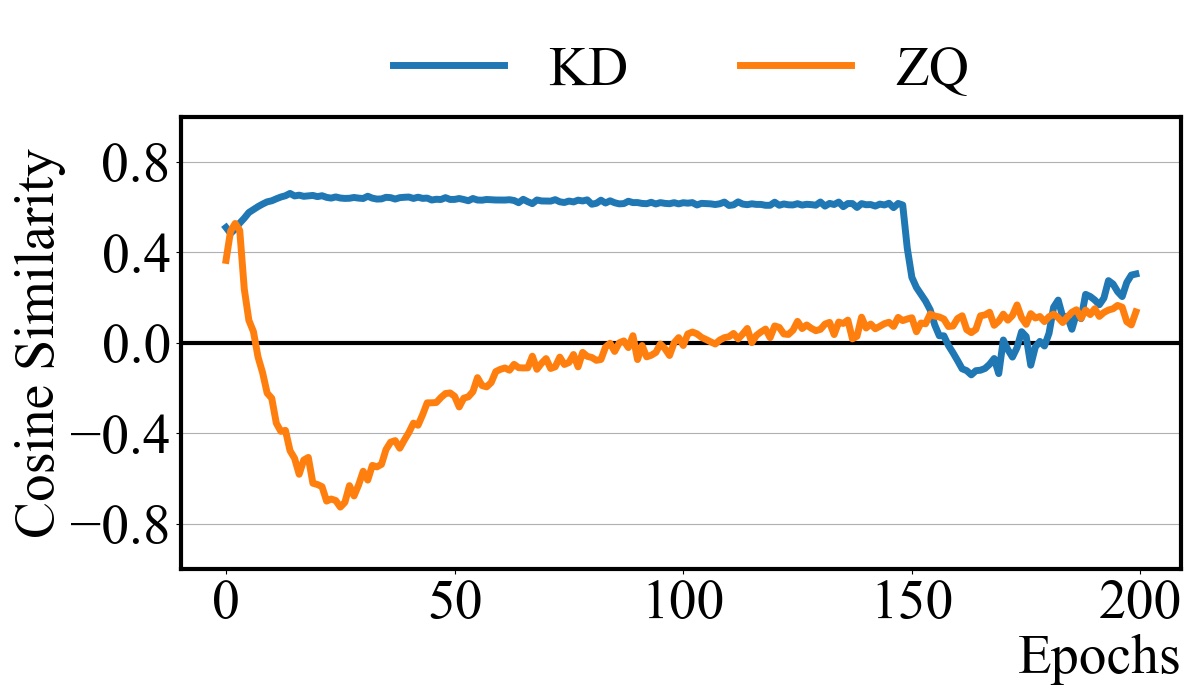}
\includegraphics[width=0.45\textwidth]{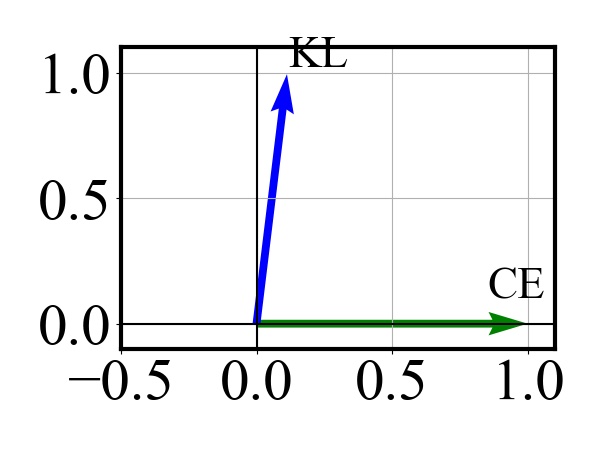}
\includegraphics[width=0.45\textwidth]{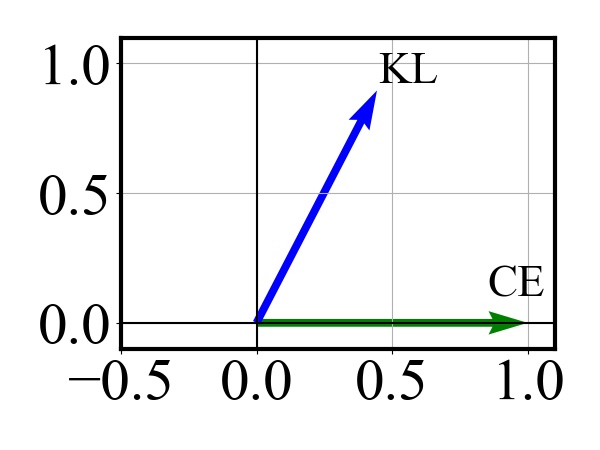}
\caption{CIFAR-10 WRN40-8}
\end{subfigure}
\caption{Gradient directionality of KL divergence and cross-entropy loss measured with CIFAR-10 dataset. 
In each setting, bottom left plots gradients under zero-shot quantization and bottom right plots gradients from knowledge distillation (self-distillation), captured from middle of the training. }
\label{fig:gradient_direction_cifar10_app}
\end{figure*}

\begin{figure*}[t]
\centering
\begin{subfigure}[t]{0.37\textwidth}
\centering
\includegraphics[width=0.9\textwidth]{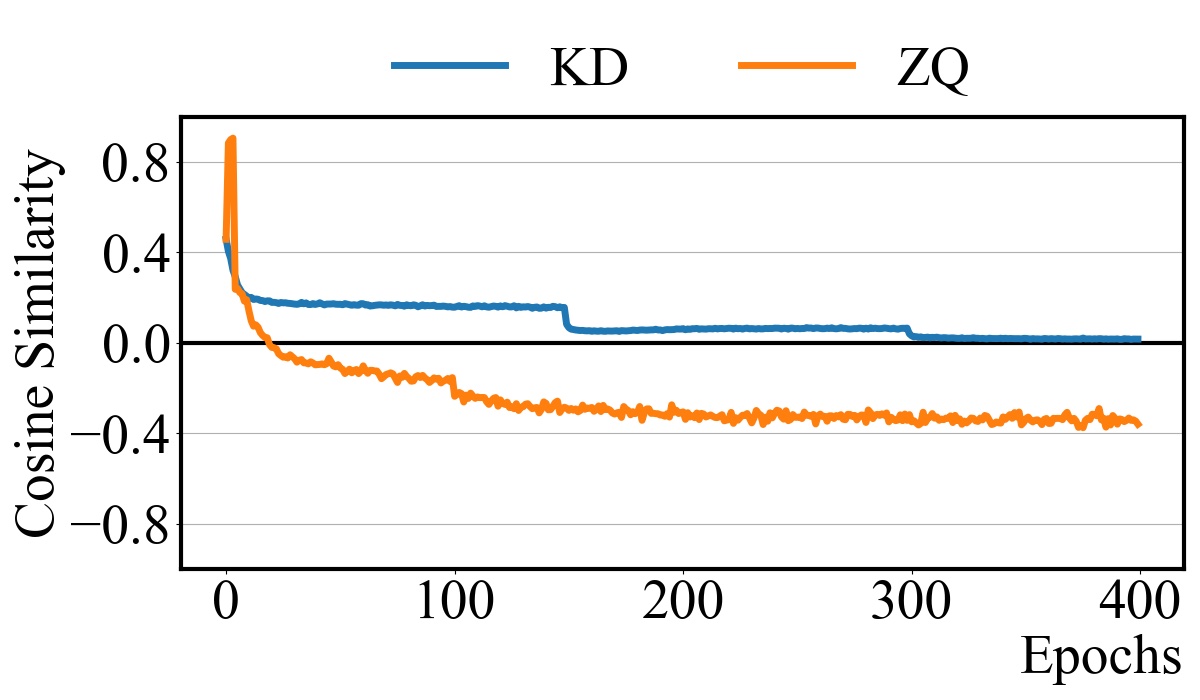}
\includegraphics[width=0.45\textwidth]{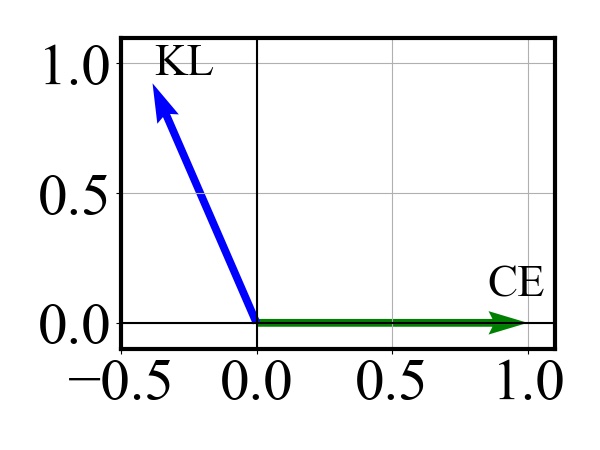}
\includegraphics[width=0.45\textwidth]{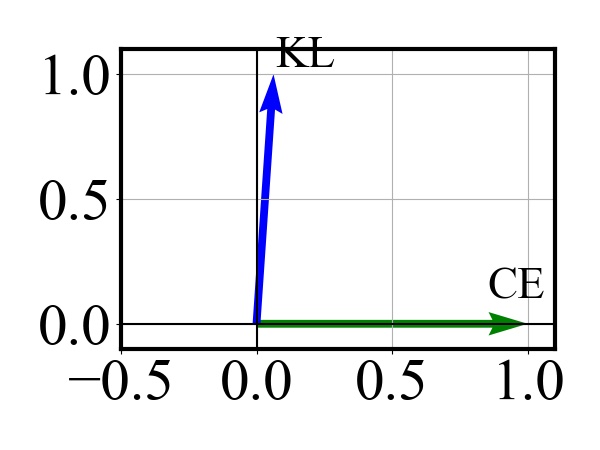}
\caption{CIFAR-100 ResNet-20}
\end{subfigure}
\hspace{10mm}
\begin{subfigure}[t]{0.37\textwidth}
\centering
\includegraphics[width=0.9\textwidth]{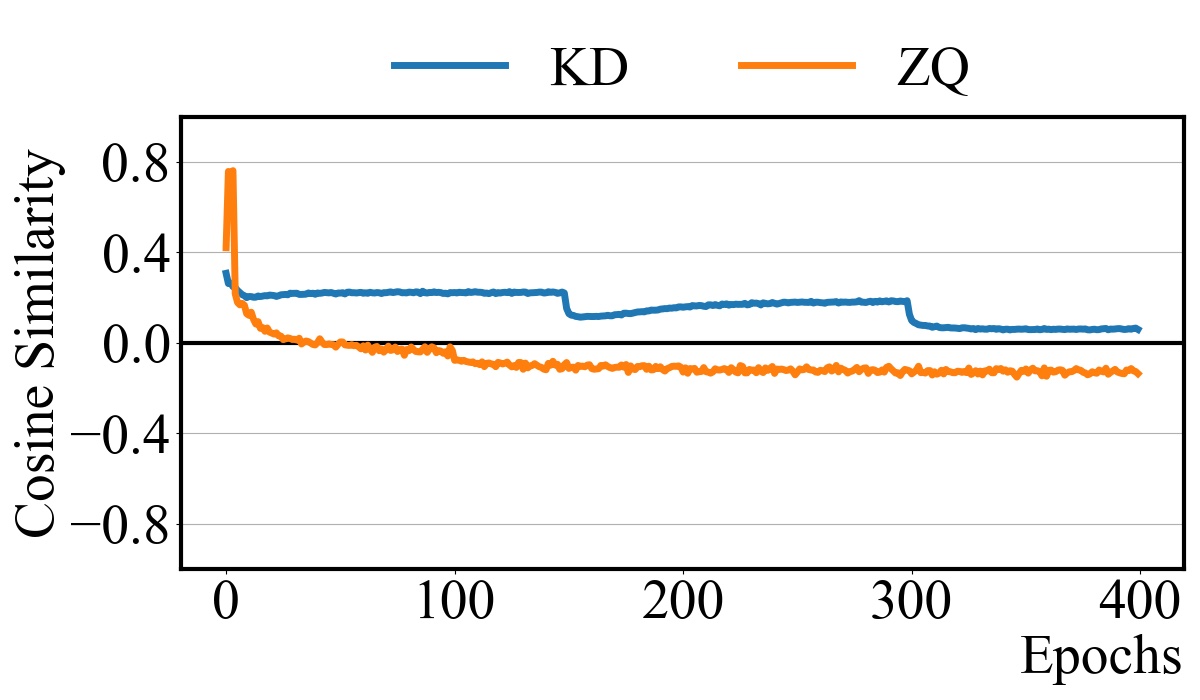}
\includegraphics[width=0.45\textwidth]{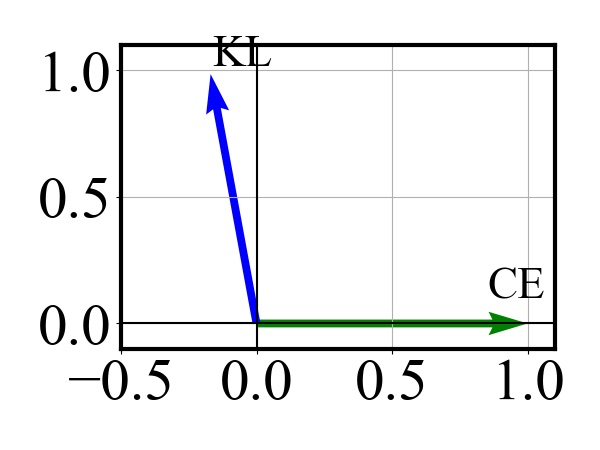}
\includegraphics[width=0.45\textwidth]{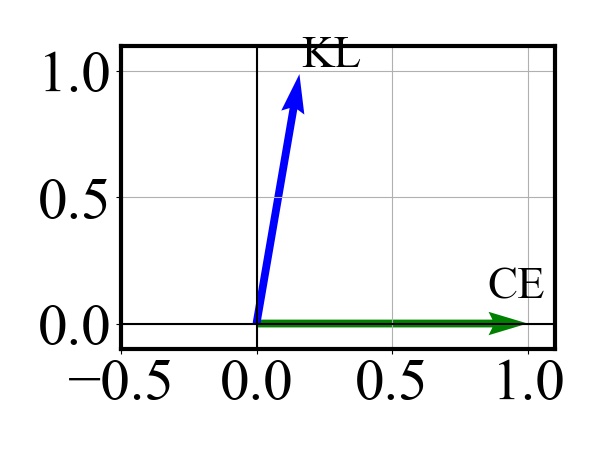}
\caption{CIFAR-100 ResNet-56}
\end{subfigure}
\begin{subfigure}[t]{0.37\textwidth}
\centering
\includegraphics[width=0.9\textwidth]{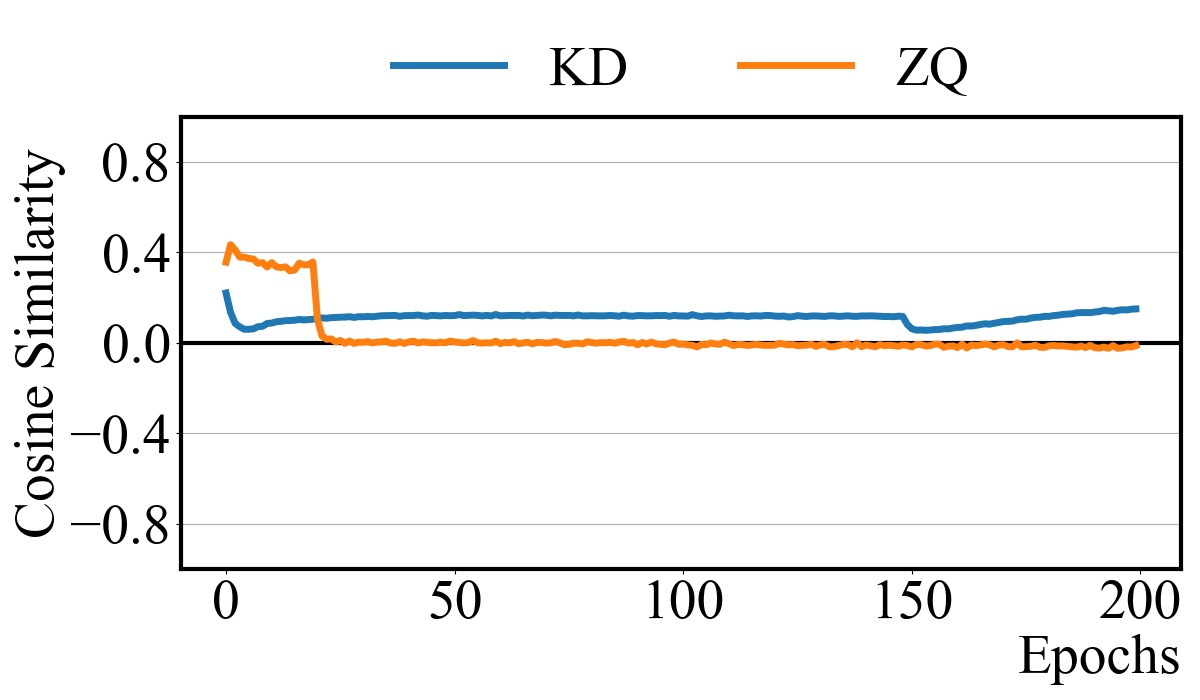}
\includegraphics[width=0.45\textwidth]{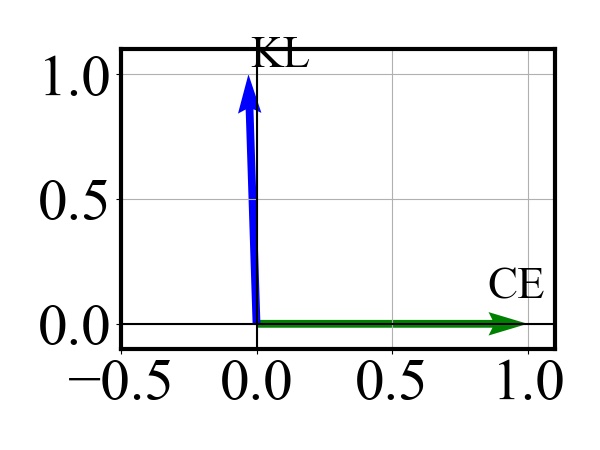}
\includegraphics[width=0.45\textwidth]{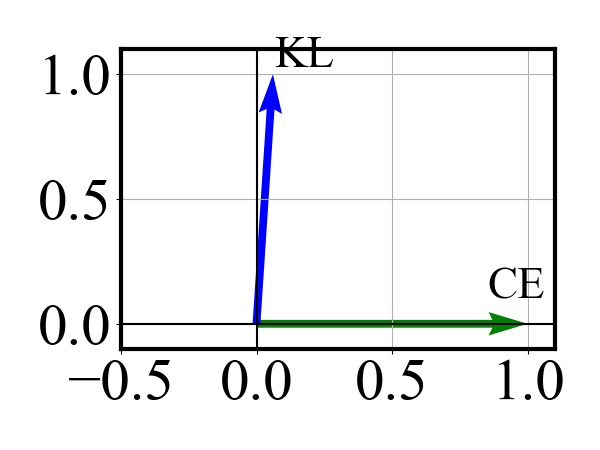}
\caption{CIFAR-100 WRN28-10}
\end{subfigure}
\hspace{10mm}
\begin{subfigure}[t]{0.37\textwidth}
\centering
\includegraphics[width=0.9\textwidth]{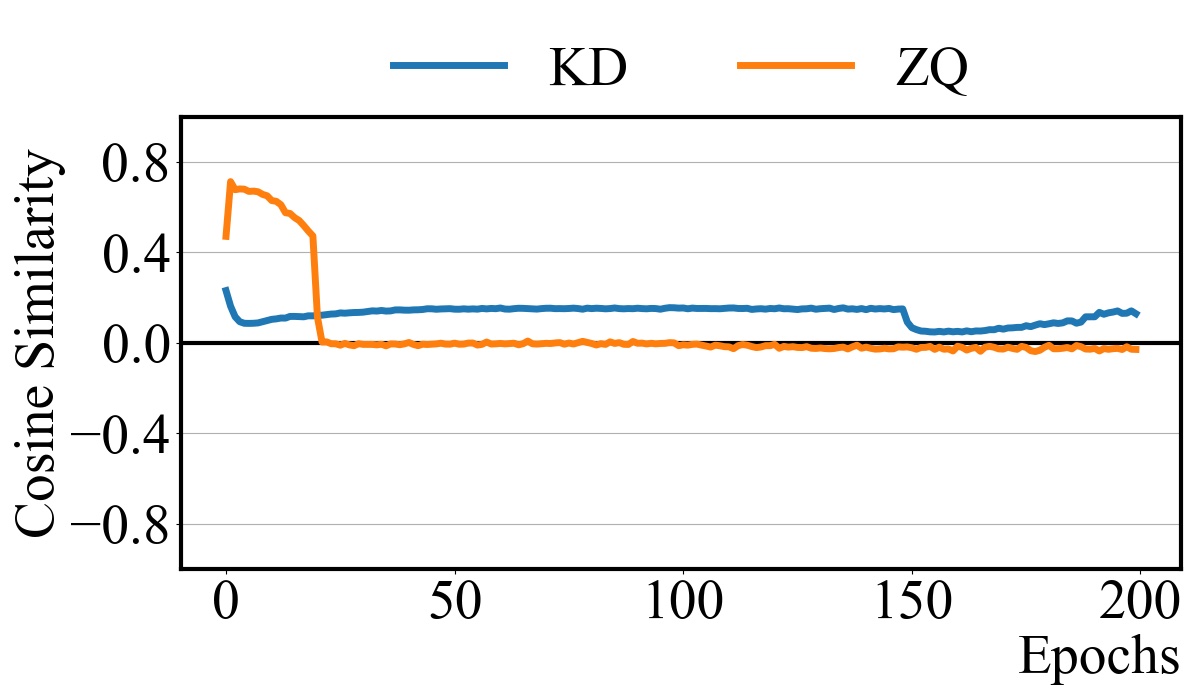}
\includegraphics[width=0.45\textwidth]{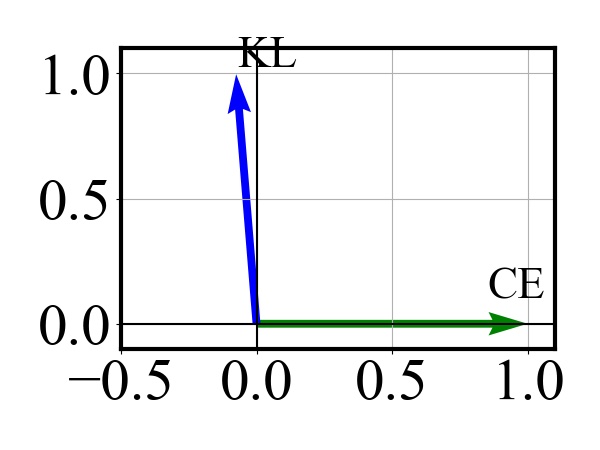}
\includegraphics[width=0.45\textwidth]{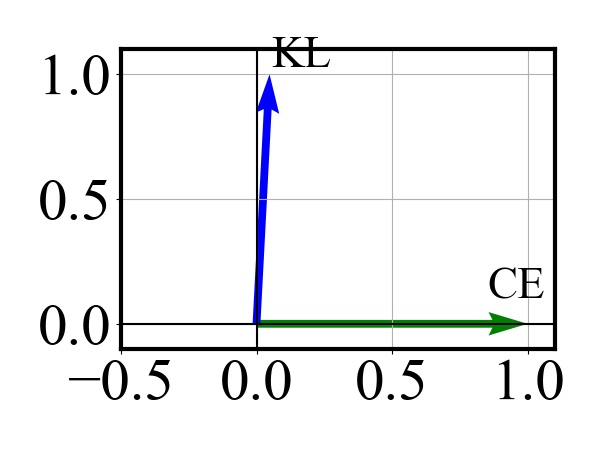}
\caption{CIFAR-100 WRN40-8}
\end{subfigure}
\caption{Gradient directionality of KL divergence and cross-entropy loss measured with CIFAR-100 dataset. 
In each setting, bottom left plots gradients under zero-shot quantization and bottom right plots gradients from knowledge distillation (self-distillation), captured from middle of the training.}
\label{fig:gradient_direction_cifar100_app}
\end{figure*}

\cref{fig:gradient_direction_cifar10_app} shows the results for CIFAR-10, and \cref{fig:gradient_direction_cifar100_app} for CIFAR-100. 
Although the quantitative difference of cosine similarities and the details of its change throughout the training differs across different datasets and networks, one trend is consistent: KL divergence and cross-entropy disagrees with each other more under the zero-shot quantization setting. 
Such tendency is usually maintained throughout the training.


\section{Hessian Trace}
In this paper, Hessian matrix was used to measure the local curvature of the loss surface and compare the generalizability of the two distinct loss terms. 
Since Hessian matrix itself is enormous in size and computations involving its entirety is considered almost infeasible, analyzing the trace value of the matrix is often the most preferred way to study its characteristics. 
Adding to our results on Section 3.2 of the main body, we share further analysis on the loss curvature using Hessian trace.

We conducted further analysis on CIFAR-10 and CIFAR-100 datasets, on four different network models: ResNet-20, ResNet-56, WRN-28, and WRN-40.
For all cases, our findings are the same.
KL divergence has much smaller local curvature than the cross-entropy, where the gap is larger in zero-shot quantization settings.

\begin{figure*}[t]
\centering
\begin{subfigure}[t]{0.45\textwidth}
    \begin{tikzpicture}

\centering
\hspace{3mm}
\begin{groupplot}[group style={vertical sep=2.8em,horizontal sep=3em,group size= 2 by 1},height=3.6cm,width=0.5\columnwidth]
\nextgroupplot[
width=.52\textwidth, 
height=3.5cm,
xmajorticks=true,
ymajorgrids,
xmin=0,
xmax=200,
ymax=15,
ymin = 0,
xlabel = {Epoch},
xlabel shift=-.3em,
xlabel near ticks,
ylabel near ticks,
xlabel style={at={(ticklabel cs:0.9)},font =\rmfamily\scriptsize},
x tick label style = {font =\scriptsize, text width = 1.4cm, align = center, anchor = north},
ylabel={Log(Tr(\textbf{H}))},
ylabel style = {font=\scriptsize, yshift=-0.2cm},
y tick label style = {font =\scriptsize, anchor = east, 
 /pgf/number format/fixed, 
},
legend cell align={left},
legend style={draw=none, fill=none, at={(0.54,1.0), font=\scriptsize},
anchor=south,legend columns=1,
/tikz/every even column/.append style={column sep=0.1cm}},
]  

\addplot[mark=none,ultra thick,color=olivegreen] table [x=Epoch, y=cifar10resnet20kdce] {figs/hessian/supp/dds.txt};\addlegendentry{CE-KD (Real Data)}

\addplot[mark=none,ultra thick,color=blue] table [x=Epoch, y=cifar10resnet20kdkl] {figs/hessian/supp/dds.txt};\addlegendentry{KL-KD (Real Data)}

\nextgroupplot[
width=.52\textwidth, 
height=3.5cm,
xmajorticks=true,
ymajorgrids,
xmin=0,
xmax=200,
ymax=15,
ymin = 0,
xlabel = {Epoch},
xlabel shift=-.3em,
xlabel near ticks,
ylabel near ticks,
xlabel style={at={(ticklabel cs:0.9)},font =\rmfamily\scriptsize},
x tick label style = {font =\scriptsize, text width = 1.4cm, align = center, anchor = north},
ylabel={Log(Tr(\textbf{H}))},
ylabel style = {font=\scriptsize, yshift=-0.2cm},
y tick label style = {font =\scriptsize, anchor = east, 
 /pgf/number format/fixed, 
},
legend cell align={left},
legend style={draw=none, fill=none, at={(0.54,1.0), font=\scriptsize},
anchor=south,legend columns=1,
/tikz/every even column/.append style={column sep=0.1cm}},
]  

\addplot[mark=none,ultra thick,color=olivegreen] table [x=Epoch, y=cifar10resnet20zqce] {figs/hessian/supp/dds.txt};\addlegendentry{CE-ZQ (Synthetic)}
\addplot[mark=none,ultra thick,color=blue] table [x=Epoch, y=cifar10resnet20zqkl] {figs/hessian/supp/dds.txt};\addlegendentry{KL-ZQ (Synthetic)}

\end{groupplot}

\end{tikzpicture}  
\caption{CIFAR-10 ResNet-20}
\end{subfigure}
\begin{subfigure}[t]{0.45\textwidth}
    \begin{tikzpicture}

\centering
\hspace{3mm}
\begin{groupplot}[group style={vertical sep=2.8em,horizontal sep=3em,group size= 2 by 1},height=3.6cm,width=0.5\columnwidth]
\nextgroupplot[
width=.52\textwidth, 
height=3.5cm,
xmajorticks=true,
ymajorgrids,
xmin=0,
xmax=200,
ymax=15,
ymin = 0,
xlabel = {Epoch},
xlabel shift=-.3em,
xlabel near ticks,
ylabel near ticks,
xlabel style={at={(ticklabel cs:0.9)},font =\rmfamily\scriptsize},
x tick label style = {font =\scriptsize, text width = 1.4cm, align = center, anchor = north},
ylabel={Log(Tr(\textbf{H}))},
ylabel style = {font=\scriptsize, yshift=-0.2cm},
y tick label style = {font =\scriptsize, anchor = east, 
 /pgf/number format/fixed, 
},
legend cell align={left},
legend style={draw=none, fill=none, at={(0.54,1.0), font=\scriptsize},
anchor=south,legend columns=1,
/tikz/every even column/.append style={column sep=0.1cm}},
]  

\addplot[mark=none,ultra thick,color=olivegreen] table [x=Epoch, y=cifar10resnet56kdce] {figs/hessian/supp/dds.txt};\addlegendentry{CE-KD (Real Data)}

\addplot[mark=none,ultra thick,color=blue] table [x=Epoch, y=cifar10resnet56kdkl] {figs/hessian/supp/dds.txt};\addlegendentry{KL-KD (Real Data)}

\nextgroupplot[
width=.52\textwidth, 
height=3.5cm,
xmajorticks=true,
ymajorgrids,
xmin=0,
xmax=200,
ymax=15,
ymin = 0,
xlabel = {Epoch},
xlabel shift=-.3em,
xlabel near ticks,
ylabel near ticks,
xlabel style={at={(ticklabel cs:0.9)},font =\rmfamily\scriptsize},
x tick label style = {font =\scriptsize, text width = 1.4cm, align = center, anchor = north},
ylabel={Log(Tr(\textbf{H}))},
ylabel style = {font=\scriptsize, yshift=-0.2cm},
y tick label style = {font =\scriptsize, anchor = east, 
 /pgf/number format/fixed, 
},
legend cell align={left},
legend style={draw=none, fill=none, at={(0.54,1.0), font=\scriptsize},
anchor=south,legend columns=1,
/tikz/every even column/.append style={column sep=0.1cm}},
]  

\addplot[mark=none,ultra thick,color=olivegreen] table [x=Epoch, y=cifar10resnet56zqce] {figs/hessian/supp/dds.txt};\addlegendentry{CE-ZQ (Synthetic)}
\addplot[mark=none,ultra thick,color=blue] table [x=Epoch, y=cifar10resnet56zqkl] {figs/hessian/supp/dds.txt};\addlegendentry{KL-ZQ (Synthetic)}

\end{groupplot}

\end{tikzpicture}  
\caption{CIFAR-10 ResNet-56}
\end{subfigure}
\begin{subfigure}[t]{0.45\textwidth}
    \begin{tikzpicture}

\centering
\hspace{3mm}
\begin{groupplot}[group style={vertical sep=2.8em,horizontal sep=3em,group size= 2 by 1},height=3.6cm,width=0.5\columnwidth]
\nextgroupplot[
width=.52\textwidth, 
height=3.5cm,
xmajorticks=true,
ymajorgrids,
xmin=0,
xmax=200,
ymax=15,
ymin = 0,
xlabel = {Epoch},
xlabel shift=-.3em,
xlabel near ticks,
ylabel near ticks,
xlabel style={at={(ticklabel cs:0.9)},font =\rmfamily\scriptsize},
x tick label style = {font =\scriptsize, text width = 1.4cm, align = center, anchor = north},
ylabel={Log(Tr(\textbf{H}))},
ylabel style = {font=\scriptsize, yshift=-0.2cm},
y tick label style = {font =\scriptsize, anchor = east, 
 /pgf/number format/fixed, 
},
legend cell align={left},
legend style={draw=none, fill=none, at={(0.54,1.0), font=\scriptsize},
anchor=south,legend columns=1,
/tikz/every even column/.append style={column sep=0.1cm}},
]  

\addplot[mark=none,ultra thick,color=olivegreen] table [x=Epoch, y=cifar10wrn28kdce] {figs/hessian/supp/dds.txt};\addlegendentry{CE-KD (Real Data)}

\addplot[mark=none,ultra thick,color=blue] table [x=Epoch, y=cifar10wrn28kdkl] {figs/hessian/supp/dds.txt};\addlegendentry{KL-KD (Real Data)}

\nextgroupplot[
width=.52\textwidth, 
height=3.5cm,
xmajorticks=true,
ymajorgrids,
xmin=0,
xmax=200,
ymax=15,
ymin = 0,
xlabel = {Epoch},
xlabel shift=-.3em,
xlabel near ticks,
ylabel near ticks,
xlabel style={at={(ticklabel cs:0.9)},font =\rmfamily\scriptsize},
x tick label style = {font =\scriptsize, text width = 1.4cm, align = center, anchor = north},
ylabel={Log(Tr(\textbf{H}))},
ylabel style = {font=\scriptsize, yshift=-0.2cm},
y tick label style = {font =\scriptsize, anchor = east, 
 /pgf/number format/fixed, 
},
legend cell align={left},
legend style={draw=none, fill=none, at={(0.54,1.0), font=\scriptsize},
anchor=south,legend columns=1,
/tikz/every even column/.append style={column sep=0.1cm}},
]  

\addplot[mark=none,ultra thick,color=olivegreen] table [x=Epoch, y=cifar10wrn28zqce] {figs/hessian/supp/dds.txt};\addlegendentry{CE-ZQ (Synthetic)}
\addplot[mark=none,ultra thick,color=blue] table [x=Epoch, y=cifar10wrn28zqkl] {figs/hessian/supp/dds.txt};\addlegendentry{KL-ZQ (Synthetic)}

\end{groupplot}

\end{tikzpicture}  
\caption{CIFAR-10 WRN-28}
\end{subfigure}
\begin{subfigure}[t]{0.45\textwidth}
    \begin{tikzpicture}

\centering
\hspace{3mm}
\begin{groupplot}[group style={vertical sep=2.8em,horizontal sep=3em,group size= 2 by 1},height=3.6cm,width=0.5\columnwidth]
\nextgroupplot[
width=.52\textwidth, 
height=3.5cm,
xmajorticks=true,
ymajorgrids,
xmin=0,
xmax=200,
ymax=15,
ymin = 0,
xlabel = {Epoch},
xlabel shift=-.3em,
xlabel near ticks,
ylabel near ticks,
xlabel style={at={(ticklabel cs:0.9)},font =\rmfamily\scriptsize},
x tick label style = {font =\scriptsize, text width = 1.4cm, align = center, anchor = north},
ylabel={Log(Tr(\textbf{H}))},
ylabel style = {font=\scriptsize, yshift=-0.2cm},
y tick label style = {font =\scriptsize, anchor = east, 
 /pgf/number format/fixed, 
},
legend cell align={left},
legend style={draw=none, fill=none, at={(0.54,1.0), font=\scriptsize},
anchor=south,legend columns=1,
/tikz/every even column/.append style={column sep=0.1cm}},
]  

\addplot[mark=none,ultra thick,color=olivegreen] table [x=Epoch, y=cifar10wrn40kdce] {figs/hessian/supp/dds.txt};\addlegendentry{CE-KD (Real Data)}

\addplot[mark=none,ultra thick,color=blue] table [x=Epoch, y=cifar10wrn40kdkl] {figs/hessian/supp/dds.txt};\addlegendentry{KL-KD (Real Data)}

\nextgroupplot[
width=.52\textwidth, 
height=3.5cm,
xmajorticks=true,
ymajorgrids,
xmin=0,
xmax=200,
ymax=15,
ymin = 0,
xlabel = {Epoch},
xlabel shift=-.3em,
xlabel near ticks,
ylabel near ticks,
xlabel style={at={(ticklabel cs:0.9)},font =\rmfamily\scriptsize},
x tick label style = {font =\scriptsize, text width = 1.4cm, align = center, anchor = north},
ylabel={Log(Tr(\textbf{H}))},
ylabel style = {font=\scriptsize, yshift=-0.2cm},
y tick label style = {font =\scriptsize, anchor = east, 
 /pgf/number format/fixed, 
},
legend cell align={left},
legend style={draw=none, fill=none, at={(0.54,1.0), font=\scriptsize},
anchor=south,legend columns=1,
/tikz/every even column/.append style={column sep=0.1cm}},
]  

\addplot[mark=none,ultra thick,color=olivegreen] table [x=Epoch, y=cifar10wrn40zqce] {figs/hessian/supp/dds.txt};\addlegendentry{CE-ZQ (Synthetic)}
\addplot[mark=none,ultra thick,color=blue] table [x=Epoch, y=cifar10wrn40zqkl] {figs/hessian/supp/dds.txt};\addlegendentry{KL-ZQ (Synthetic)}

\end{groupplot}

\end{tikzpicture}  
\caption{CIFAR-10 WRN-40}
\end{subfigure}
\begin{subfigure}[t]{0.45\textwidth}
    \begin{tikzpicture}

\centering
\hspace{3mm}
\begin{groupplot}[group style={vertical sep=2.8em,horizontal sep=3em,group size= 2 by 1},height=3.6cm,width=0.5\columnwidth]
\nextgroupplot[
width=.52\textwidth, 
height=3.5cm,
xmajorticks=true,
ymajorgrids,
xmin=0,
xmax=200,
ymax=15,
ymin = 0,
xlabel = {Epoch},
xlabel shift=-.3em,
xlabel near ticks,
ylabel near ticks,
xlabel style={at={(ticklabel cs:0.9)},font =\rmfamily\scriptsize},
x tick label style = {font =\scriptsize, text width = 1.4cm, align = center, anchor = north},
ylabel={Log(Tr(\textbf{H}))},
ylabel style = {font=\scriptsize, yshift=-0.2cm},
y tick label style = {font =\scriptsize, anchor = east, 
 /pgf/number format/fixed, 
},
legend cell align={left},
legend style={draw=none, fill=none, at={(0.54,1.0), font=\scriptsize},
anchor=south,legend columns=1,
/tikz/every even column/.append style={column sep=0.1cm}},
]  

\addplot[mark=none,ultra thick,color=olivegreen] table [x=Epoch, y=cifar100resnet20kdce] {figs/hessian/supp/dds.txt};\addlegendentry{CE-KD (Real Data)}

\addplot[mark=none,ultra thick,color=blue] table [x=Epoch, y=cifar100resnet20kdkl] {figs/hessian/supp/dds.txt};\addlegendentry{KL-KD (Real Data)}

\nextgroupplot[
width=.52\textwidth, 
height=3.5cm,
xmajorticks=true,
ymajorgrids,
xmin=0,
xmax=200,
ymax=15,
ymin = 0,
xlabel = {Epoch},
xlabel shift=-.3em,
xlabel near ticks,
ylabel near ticks,
xlabel style={at={(ticklabel cs:0.9)},font =\rmfamily\scriptsize},
x tick label style = {font =\scriptsize, text width = 1.4cm, align = center, anchor = north},
ylabel={Log(Tr(\textbf{H}))},
ylabel style = {font=\scriptsize, yshift=-0.2cm},
y tick label style = {font =\scriptsize, anchor = east, 
 /pgf/number format/fixed, 
},
legend cell align={left},
legend style={draw=none, fill=none, at={(0.54,1.0), font=\scriptsize},
anchor=south,legend columns=1,
/tikz/every even column/.append style={column sep=0.1cm}},
]  

\addplot[mark=none,ultra thick,color=olivegreen] table [x=Epoch, y=cifar100resnet20zqce] {figs/hessian/supp/dds.txt};\addlegendentry{CE-ZQ (Synthetic)}
\addplot[mark=none,ultra thick,color=blue] table [x=Epoch, y=cifar100resnet20zqkl] {figs/hessian/supp/dds.txt};\addlegendentry{KL-ZQ (Synthetic)}

\end{groupplot}

\end{tikzpicture}  
\caption{CIFAR-100 ResNet-20}
\end{subfigure}
\begin{subfigure}[t]{0.45\textwidth}
    \begin{tikzpicture}

\centering
\hspace{3mm}
\begin{groupplot}[group style={vertical sep=2.8em,horizontal sep=3em,group size= 2 by 1},height=3.6cm,width=0.5\columnwidth]
\nextgroupplot[
width=.52\textwidth, 
height=3.5cm,
xmajorticks=true,
ymajorgrids,
xmin=0,
xmax=200,
ymax=15,
ymin = 0,
xlabel = {Epoch},
xlabel shift=-.3em,
xlabel near ticks,
ylabel near ticks,
xlabel style={at={(ticklabel cs:0.9)},font =\rmfamily\scriptsize},
x tick label style = {font =\scriptsize, text width = 1.4cm, align = center, anchor = north},
ylabel={Log(Tr(\textbf{H}))},
ylabel style = {font=\scriptsize, yshift=-0.2cm},
y tick label style = {font =\scriptsize, anchor = east, 
 /pgf/number format/fixed, 
},
legend cell align={left},
legend style={draw=none, fill=none, at={(0.54,1.0), font=\scriptsize},
anchor=south,legend columns=1,
/tikz/every even column/.append style={column sep=0.1cm}},
]  

\addplot[mark=none,ultra thick,color=olivegreen] table [x=Epoch, y=cifar100resnet56kdce] {figs/hessian/supp/dds.txt};\addlegendentry{CE-KD (Real Data)}

\addplot[mark=none,ultra thick,color=blue] table [x=Epoch, y=cifar100resnet56kdkl] {figs/hessian/supp/dds.txt};\addlegendentry{KL-KD (Real Data)}

\nextgroupplot[
width=.52\textwidth, 
height=3.5cm,
xmajorticks=true,
ymajorgrids,
xmin=0,
xmax=200,
ymax=15,
ymin = 0,
xlabel = {Epoch},
xlabel shift=-.3em,
xlabel near ticks,
ylabel near ticks,
xlabel style={at={(ticklabel cs:0.9)},font =\rmfamily\scriptsize},
x tick label style = {font =\scriptsize, text width = 1.4cm, align = center, anchor = north},
ylabel={Log(Tr(\textbf{H}))},
ylabel style = {font=\scriptsize, yshift=-0.2cm},
y tick label style = {font =\scriptsize, anchor = east, 
 /pgf/number format/fixed, 
},
legend cell align={left},
legend style={draw=none, fill=none, at={(0.54,1.0), font=\scriptsize},
anchor=south,legend columns=1,
/tikz/every even column/.append style={column sep=0.1cm}},
]  

\addplot[mark=none,ultra thick,color=olivegreen] table [x=Epoch, y=cifar100resnet56zqce] {figs/hessian/supp/dds.txt};\addlegendentry{CE-ZQ (Synthetic)}
\addplot[mark=none,ultra thick,color=blue] table [x=Epoch, y=cifar100resnet56zqkl] {figs/hessian/supp/dds.txt};\addlegendentry{KL-ZQ (Synthetic)}

\end{groupplot}

\end{tikzpicture}  
\caption{CIFAR-100 ResNet-56}
\end{subfigure}
\begin{subfigure}[t]{0.45\textwidth}
    \begin{tikzpicture}

\centering
\hspace{3mm}
\begin{groupplot}[group style={vertical sep=2.8em,horizontal sep=3em,group size= 2 by 1},height=3.6cm,width=0.5\columnwidth]
\nextgroupplot[
width=.52\textwidth, 
height=3.5cm,
xmajorticks=true,
ymajorgrids,
xmin=0,
xmax=200,
ymax=15,
ymin = 0,
xlabel = {Epoch},
xlabel shift=-.3em,
xlabel near ticks,
ylabel near ticks,
xlabel style={at={(ticklabel cs:0.9)},font =\rmfamily\scriptsize},
x tick label style = {font =\scriptsize, text width = 1.4cm, align = center, anchor = north},
ylabel={Log(Tr(\textbf{H}))},
ylabel style = {font=\scriptsize, yshift=-0.2cm},
y tick label style = {font =\scriptsize, anchor = east, 
 /pgf/number format/fixed, 
},
legend cell align={left},
legend style={draw=none, fill=none, at={(0.54,1.0), font=\scriptsize},
anchor=south,legend columns=1,
/tikz/every even column/.append style={column sep=0.1cm}},
]  

\addplot[mark=none,ultra thick,color=olivegreen] table [x=Epoch, y=cifar100wrn28kdce] {figs/hessian/supp/dds.txt};\addlegendentry{CE-KD (Real Data)}

\addplot[mark=none,ultra thick,color=blue] table [x=Epoch, y=cifar100wrn28kdkl] {figs/hessian/supp/dds.txt};\addlegendentry{KL-KD (Real Data)}

\nextgroupplot[
width=.52\textwidth, 
height=3.5cm,
xmajorticks=true,
ymajorgrids,
xmin=0,
xmax=200,
ymax=15,
ymin = 0,
xlabel = {Epoch},
xlabel shift=-.3em,
xlabel near ticks,
ylabel near ticks,
xlabel style={at={(ticklabel cs:0.9)},font =\rmfamily\scriptsize},
x tick label style = {font =\scriptsize, text width = 1.4cm, align = center, anchor = north},
ylabel={Log(Tr(\textbf{H}))},
ylabel style = {font=\scriptsize, yshift=-0.2cm},
y tick label style = {font =\scriptsize, anchor = east, 
 /pgf/number format/fixed, 
},
legend cell align={left},
legend style={draw=none, fill=none, at={(0.54,1.0), font=\scriptsize},
anchor=south,legend columns=1,
/tikz/every even column/.append style={column sep=0.1cm}},
]  

\addplot[mark=none,ultra thick,color=olivegreen] table [x=Epoch, y=cifar100wrn28zqce] {figs/hessian/supp/dds.txt};\addlegendentry{CE-ZQ (Synthetic)}
\addplot[mark=none,ultra thick,color=blue] table [x=Epoch, y=cifar100wrn28zqkl] {figs/hessian/supp/dds.txt};\addlegendentry{KL-ZQ (Synthetic)}

\end{groupplot}

\end{tikzpicture}  
\caption{CIFAR-100 WRN-28}
\end{subfigure}
\begin{subfigure}[t]{0.45\textwidth}
    \begin{tikzpicture}

\centering
\hspace{3mm}
\begin{groupplot}[group style={vertical sep=2.8em,horizontal sep=3em,group size= 2 by 1},height=3.6cm,width=0.5\columnwidth]
\nextgroupplot[
width=.52\textwidth, 
height=3.5cm,
xmajorticks=true,
ymajorgrids,
xmin=0,
xmax=200,
ymax=15,
ymin = 0,
xlabel = {Epoch},
xlabel shift=-.3em,
xlabel near ticks,
ylabel near ticks,
xlabel style={at={(ticklabel cs:0.9)},font =\rmfamily\scriptsize},
x tick label style = {font =\scriptsize, text width = 1.4cm, align = center, anchor = north},
ylabel={Log(Tr(\textbf{H}))},
ylabel style = {font=\scriptsize, yshift=-0.2cm},
y tick label style = {font =\scriptsize, anchor = east, 
 /pgf/number format/fixed, 
},
legend cell align={left},
legend style={draw=none, fill=none, at={(0.54,1.0), font=\scriptsize},
anchor=south,legend columns=1,
/tikz/every even column/.append style={column sep=0.1cm}},
]  

\addplot[mark=none,ultra thick,color=olivegreen] table [x=Epoch, y=cifar100wrn40kdce] {figs/hessian/supp/dds.txt};\addlegendentry{CE-KD (Real Data)}

\addplot[mark=none,ultra thick,color=blue] table [x=Epoch, y=cifar100wrn40kdkl] {figs/hessian/supp/dds.txt};\addlegendentry{KL-KD (Real Data)}

\nextgroupplot[
width=.52\textwidth, 
height=3.5cm,
xmajorticks=true,
ymajorgrids,
xmin=0,
xmax=200,
ymax=15,
ymin = 0,
xlabel = {Epoch},
xlabel shift=-.3em,
xlabel near ticks,
ylabel near ticks,
xlabel style={at={(ticklabel cs:0.9)},font =\rmfamily\scriptsize},
x tick label style = {font =\scriptsize, text width = 1.4cm, align = center, anchor = north},
ylabel={Log(Tr(\textbf{H}))},
ylabel style = {font=\scriptsize, yshift=-0.2cm},
y tick label style = {font =\scriptsize, anchor = east, 
 /pgf/number format/fixed, 
},
legend cell align={left},
legend style={draw=none, fill=none, at={(0.54,1.0), font=\scriptsize},
anchor=south,legend columns=1,
/tikz/every even column/.append style={column sep=0.1cm}},
]  

\addplot[mark=none,ultra thick,color=olivegreen] table [x=Epoch, y=cifar100wrn40zqce] {figs/hessian/supp/dds.txt};\addlegendentry{CE-ZQ (Synthetic)}
\addplot[mark=none,ultra thick,color=blue] table [x=Epoch, y=cifar100wrn40zqkl] {figs/hessian/supp/dds.txt};\addlegendentry{KL-ZQ (Synthetic)}

\end{groupplot}

\end{tikzpicture}  
\caption{CIFAR-100 WRN-40}
\end{subfigure}
\caption{Hessian trace of KL divergence and cross-entropy, measured across diverse datasets and networks.}
\end{figure*}
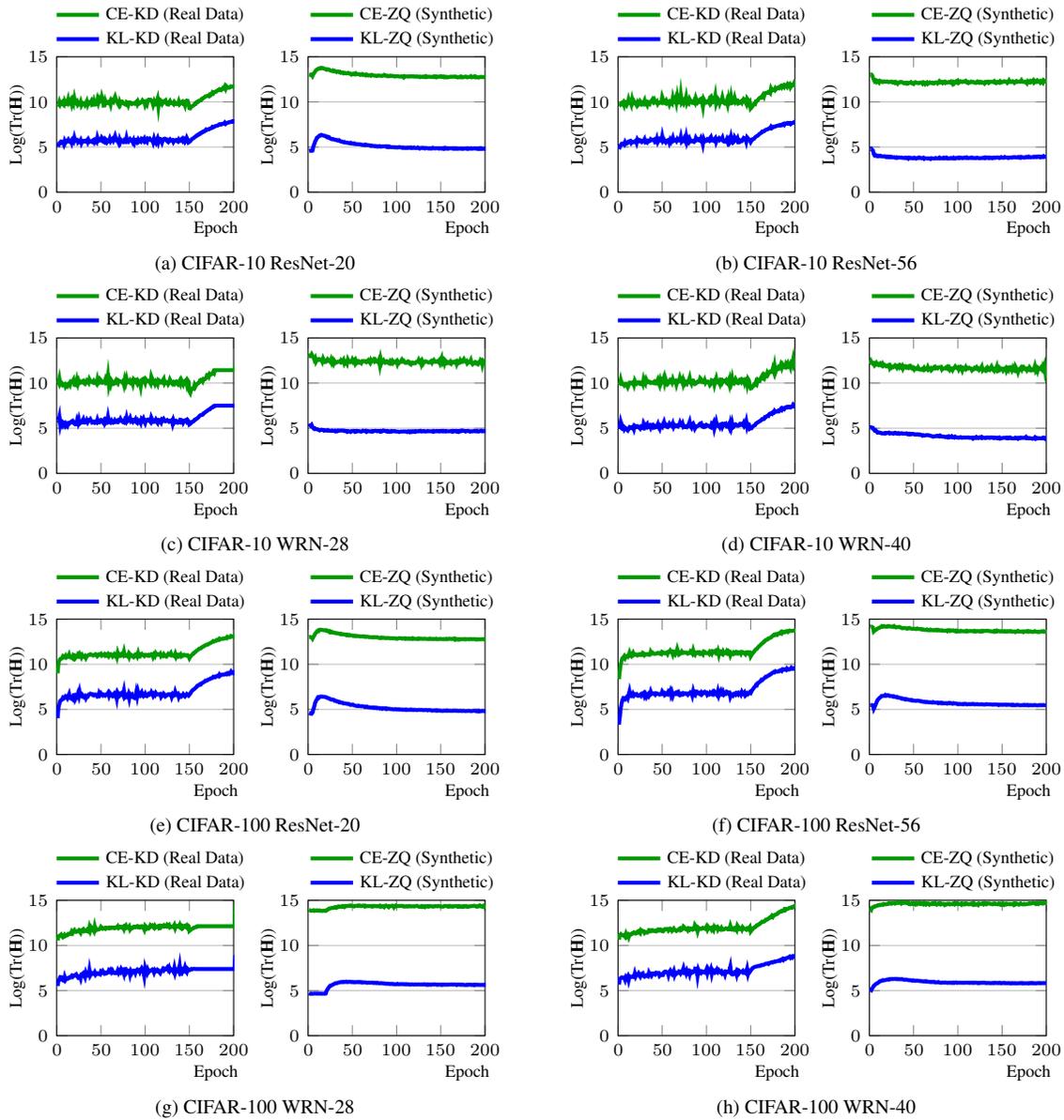

{\small
\bibliographystyleApp{ieee_fullname}
\bibliographyApp{egbib_app}
}

\end{document}